\documentclass{article} %
\usepackage[preprint]{colm2025_conference}

\usepackage{microtype}
\usepackage{hyperref}
\usepackage{url}
\usepackage{booktabs}

\definecolor{iccvblue}{rgb}{0.21,0.49,0.74}
\definecolor{figurepurple}{rgb}{0.847, 0.431, 0.8}
\usepackage{graphicx}
\usepackage{wrapfig}
\usepackage{amsmath}
\usepackage{amssymb}
\usepackage{circledsteps}
\usepackage{lipsum}
\usepackage{lineno}
\usepackage{float}

\newcommand{\boldone}[1]{\textbf{\textcolor{black}{#1}}}
\newcommand{\boldtwo}[1]{\textbf{\textcolor{blue}{#1}}}
\newcommand{\boldthree}[1]{\textbf{\textcolor{purple}{#1}}}

\definecolor{darkblue}{rgb}{0, 0, 0.5}
\hypersetup{colorlinks=true, citecolor=darkblue, linkcolor=darkblue, urlcolor=darkblue}

\title{Hidden in plain sight: \\VLMs overlook their visual representations}

\author{Stephanie Fu \\
UC Berkeley\\
\And
Tyler Bonnen \\
UC Berkeley \\
\And
Devin Guillory \\
UC Berkeley \\
\And
Trevor Darrell \\
UC Berkeley \\
}

\begin{document}

\ifcolmsubmission
\linenumbers
\fi

\maketitle

\vspace{-6mm}
\begin{figure}[h]
    \includegraphics[width=\textwidth]{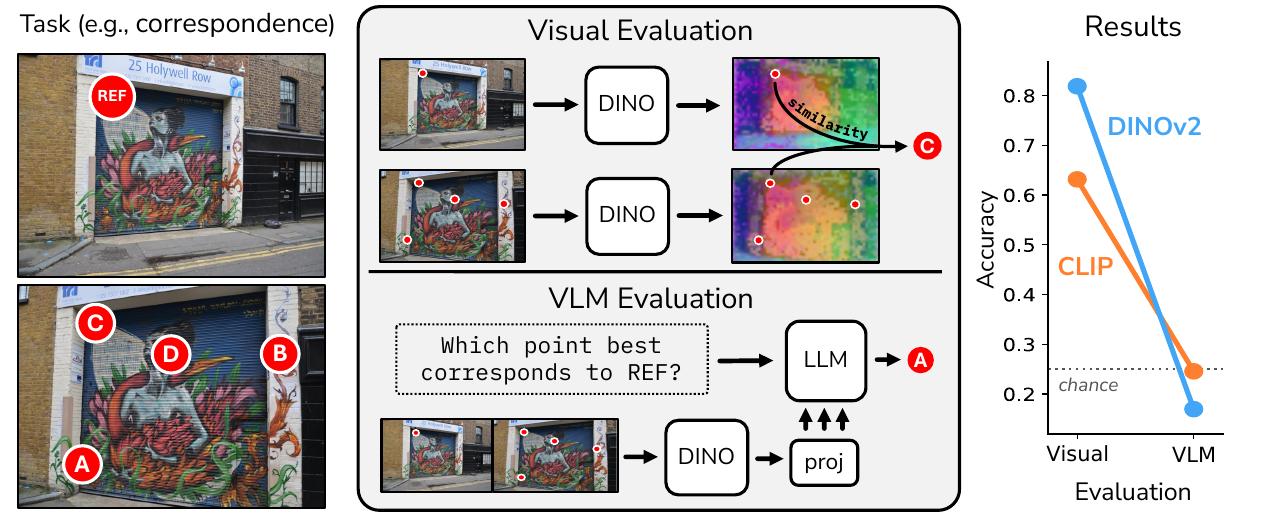}
    \vspace{-4mm}
    \caption{\label{fig:teaser} %
    \small \textbf{Evaluating vision language models (VLMs) alongside their vision encoders reveals a failure to utilize visual information.} 
    To assess VLMs' visual abilities, we compare their performance to the accuracy supported by a direct readout of their visual encoders. 
    Using `vision-centric' tasks (e.g., visual correspondence), we compare typical VQA-style VLM evaluation (center, bottom) with vision-only methods (center, top). Across tasks, performance plummets from the `Visual' to `VLM' evaluations, often from near-ceiling to random chance. We study this trend by analyzing vision representation quality, prompt sensitivity, and the LLM's ability to leverage visual information.
    }
\end{figure}

\begin{abstract}

   \vspace{-1mm}
   \noindent %
   Language provides a natural interface to specify and evaluate performance on visual tasks. 
   To realize this possibility, vision language models (VLMs) must successfully integrate visual and linguistic information.
  Our work compares VLMs to a direct readout of their visual encoders to understand their ability to integrate across these modalities. Across a series of vision-centric benchmarks (e.g., depth estimation, correspondence), we find that VLMs perform substantially worse than their visual encoders, dropping to near-chance performance. We investigate these results through a series of analyses across the entire VLM: namely 1) the degradation of vision representations, 2) brittleness to task prompt, and 3) the language model's role in solving the task. We find that the bottleneck in performing these vision-centric tasks lies in this third category; VLMs are not effectively using visual information easily accessible throughout the \textit{entire} model, and they inherit the language priors present in the LLM. 
   Our work helps diagnose the failure modes of open-source VLMs, and presents a series of evaluations useful for future investigations into visual understanding within VLMs.

\end{abstract}

\vspace{-4mm}
\section{Introduction}
\label{sec:intro}

\begin{figure*}[t]
    \centering
    \includegraphics[width=1\linewidth]{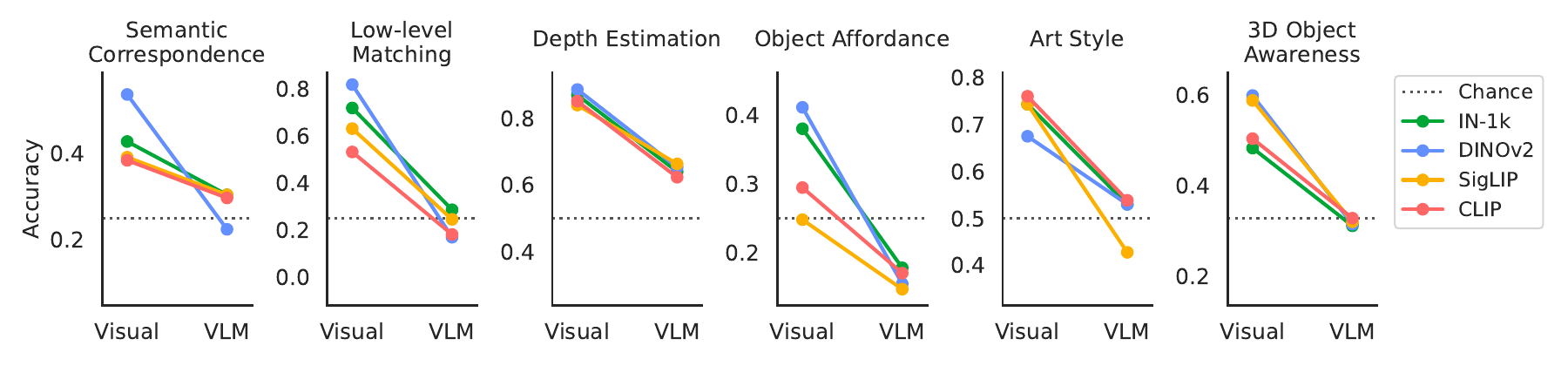}
    \vspace{-6mm}
    \caption{\small \textbf{Comparing standard visual evaluation to VLMs across vision-centric tasks.} Shifting from a standard vision evaluation strategy to a VLM evaluation results in a performance drop, often to chance-level accuracies. Additionally, the vision encoders that perform best at a task (often DINOv2) are not the same vision encoders in more performant VLMs. 
    \vspace{-5mm}
    }
    \label{fig:main_results}
\end{figure*}
Vision language models (VLMs) are designed to harness the power of both large language models (LLMs) and large-scale vision encoders to perform multimodal tasks. 
Open-source VLMs often comprise three main components - a pretrained vision encoder, projector, and LLM - and offer a flexible interface to specify visual tasks simply through a text prompt. It has also been suggested that VLMs also provide a novel paradigm for both understanding the representations of vision models and leveraging them for downstream applications \cite{tongCambrian1FullyOpen2024}. Nonetheless, the behavior of VLMs in vision-centric tasks reveals serious issues with the integration of vision and language in these models, and warrants caution in drawing conclusions about their underlying vision representations.

Open-source VLMs excel at many tasks, especially ones reliant on expert-level knowledge within the LLM. However, when VLMs need to rely solely on the image and not on background knowledge, they exhibit much worse, often random chance-level, performance \cite{tongCambrian1FullyOpen2024, fuBLINKMultimodalLarge2024}. 
These results are especially puzzling when compared with prior work reporting that large vision models - the same ones embedded in VLMs - excel on many of these `vision-centric tasks' when probing their representations \cite{bananiProbing3DAwareness2024}. 

Additionally, we observe no correlation in performance between vision encoders and their analogous VLMs on vision-centric tasks; the strongest vision representations (often DINOv2 \cite{oquabDINOv2LearningRobust2024}) tend to underperform when used within a VLM \cite{tongCambrian1FullyOpen2024, bananiProbing3DAwareness2024}. We find that these two observations - the consistent accuracy drop and unreliable rank-ordering of vision model performance - are prevalent even in state-of-the-art open-source VLMs. Why do we observe such a disconnect between the strength of vision representations and their contributions to VLM capabilities?

To better understand this phenomenon, we compare VLMs to their visual encoders and diagnose these discrepancies.
We define our testbed to be `vision-centric' tasks: those solvable purely from visual input, independent of domain expertise. This removes reliance on LLM knowledge and focuses more on the vision representations.
We compare VLMs to their visual backbones across diverse vision-centric tasks from CV-Bench \cite{tongCambrian1FullyOpen2024}, BLINK \cite{fuBLINKMultimodalLarge2024}, and MOCHI \cite{bonnen2024evaluatingmultiviewobjectconsistency}. These tasks span low- to high-level visual abilities, single/multi-image inputs, and different visual prompting strategies. We analyze three potential failure points: vision representation quality, sensitivity to the prompt, and the language model's ability to use visual representations. Our key findings are:

\begin{enumerate} 
    \item Shifting from standard visual probing strategies to a VLM-based evaluation results in a universal drop in performance, often to random chance. This trend persists across pretraining strategies for vision encoders, and does not preserve the rank-ordering of models on vision-centric benchmarks.
    \begin{itemize}
        \item Additionally, vision representations throughout projector and LLM layers do not degrade and can still solve the task with a visual probing strategy.
    \end{itemize}
    \item Prompt-tuning the VLM improves performance marginally, but with diminishing improvements and does not close the gap with standard vision evaluation strategies.
    \item The LLM's ability to use its vision representations is a limiting factor in VLM performance. Across most tasks, finetuning the LLM results in a higher accuracy than finetuning the projector or vision encoder. 
    \begin{itemize}
        \item VLM answer distributions largely reflect their blind answers. Finetuning the LLM (as opposed to the projector or ViT) also best overcomes these biases.
    \end{itemize}
\end{enumerate}

\noindent Our work demonstrate that VLMs fail to utilize representations \textit{easily-accessible} within their visual backbones (and throughout the whole VLM), and that improvements to the LLM's ability to use them offers the most promising returns for vision-centric capabilities.

\begin{figure*}[t]
    \centering
    \includegraphics[width=1\linewidth]{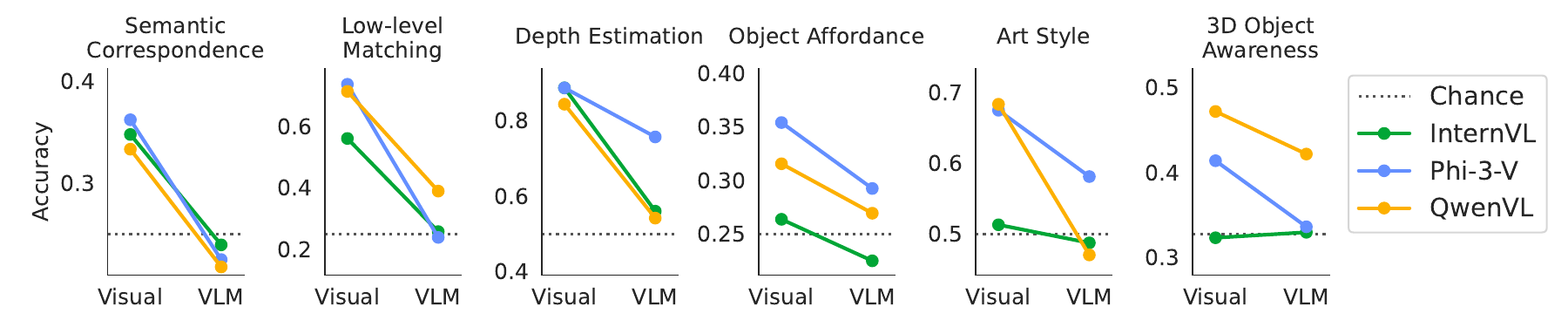}
    \vspace{-6mm}
    \caption{\small We find the same trends as in Fig. \ref{fig:main_results} for common open-source VLMs. We also note that these VLMs instruction-tune their vision encoders along with the rest of the VLM, so they are designed to be most performant when used in tandem with their projector and LLM. Nevertheless, we still see higher task performance when probing the vision representations alone than when querying the VLM.
    \vspace{-4mm}
    }
    \label{fig:sota_results}
\end{figure*}

\section{Evaluation Setup}
\label{sec:methods}
\vspace{-2mm}

To study the effect of evaluation strategy on task performance, we evaluate 4 different vision backbones: DINOv2 L/14 (self-supervised vision-only pretraining) \cite{dosovitskiyImageWorth16x162021, steiner2022trainvitdataaugmentation}, ViT-IN1k L/16 (supervised vision-only pretraining on ImageNet \cite{5206848}), and CLIP L/14 and SigLIP L/14 (both vision-language pretraining). We use checkpoints from \cite{karamchetiPrismaticVLMsInvestigating2024}, which pairs each vision backbone with Vicuña v1.5 \cite{vicuna2023}. The projection layers are trained for vision-language alignment with the LLaVA v1.5 data mixture \cite{liuImprovedBaselinesVisual2023}, with the ViT and LLM components kept frozen. As a result, the vision encoder weights in each VLM are identical to the original ViTs, allowing us to properly ablate the use of an LLM decoder while keeping the vision part constant. However, to demonstrate that these trends generalize, we also evaluate on more widely-used VLMs: Qwen-VL \cite{bai2023qwenvlversatilevisionlanguagemodel}, Phi-3-V \cite{abdin2024phi3technicalreporthighly}, and InternVL \cite{chen2024internvl}. These models also use ViTs, but their vision encoders are especially trained to be used in that VLM context.

We pick `vision-centric' tasks - that is, tasks where the model's ability to discern visual characteristics of its inputs and does not rely on language-level knowledge or domain expertise. These vision-centric abilities stem from the expressivity of the vision backbone itself, and we frame all other components of our evaluation as a method to evaluate the vision representations. 
See the Supplement for full evaluation details and task examples.

\vspace{-2mm}
\subsection{Depth Estimation} 
We evaluate depth estimation with the Depth Order task from CV-Bench \cite{tongCambrian1FullyOpen2024}. In each example, we present an image with two bounding boxes around objects and ask which is closer to the camera. 
To evaluate the vision encoder, we train a DPT head following \cite{bananiProbing3DAwareness2024, ranftlVisionTransformersDense2021} on the NYUv2 training set \cite{Silberman:ECCV12} for 10 epochs. We then crop the predicted depth map to each bounding box and compare average depth in each box to determine the object closer to the camera.
\vspace{-1mm}
\subsection{Correspondence-based Evaluations}
\vspace{-1mm}
The following three tasks, drawn from \cite{fuBLINKMultimodalLarge2024}, test for different forms of pixel-matching. Given a reference image with a dot and a second image with four dots (labeled `A', `B', `C', and `D'), the model must pick the option corresponding to the reference dot. To evaluate the vision encoder, we compute cosine similarity between the patch features of the reference and the four options, and choose the letter option with the highest similarity score.
\vspace{-5mm}

\paragraph{Semantic Correspondence.} 

We evaluate semantic correspondence with the SPair-71k dataset 
\cite{min2019spair71klargescalebenchmarksemantic} and keypoints from BLINK. This data tests for the model's ability to match object parts that visually look similar despite large variation within the object class (e.g., left leg of two different animals, fronts of two different cars in different poses). 

\vspace{-3mm}
\paragraph{Object Affordance.} 

We use BLINK's Functional Correspondence benchmark, which uses the FunKPoint dataset \cite{lai2021functional}. Each task consists of an image pair whose match shares an object function (e.g., `handle') with the reference point. Unlike Semantic Correspondence, each example does not necessarily present objects from the same class and instead focuses on the shared purpose of two object parts.

\vspace{-3mm}
\paragraph{Low-level Matching.} 

In addition to evaluating the ability to match pixels based on function or semantics, we want to understand how LLM decoding affects low-level pattern matching.
We use BLINK's benchmark here \cite{fuBLINKMultimodalLarge2024}, which draws on HPatches \cite{hpatches_2017_cvpr} containing image pairs of the same scene differing in illumination or viewpoint. The ground truth is computed with the homographies provided in HPatches.

\subsection{3D Object Awareness} 
\vspace{-1mm}

We examine not only model adherence to physical truths but also alignment with complex human visual judgments. To this end, we test for object-centric representations with the MOCHI \cite{bonnen2024evaluatingmultiviewobjectconsistency} benchmark, where participants and models identify the image containing a differnet object among 3 or 4 options. We evaluate VLMs through a multiple-choice VQA format, and evaluate the vision encoder by computing pairwise cosine similarity of CLS embeddings and choosing the example with the lowest average score.

\subsection{Art Style}
\vspace{-1mm}
We evaluate models on the Art Style benchmark from \cite{fuBLINKMultimodalLarge2024}, which sources data from WikiArt. Unlike previous VLM benchmarks on art style requiring domain knowledge \cite{yue2024mmmumassivemultidisciplinemultimodal}, BLINK's benchmark measures a model's ability to match visual characteristics without knowledge about artist name or historical context. Given a reference image, the model selects which of two image options better matches its style. 

To evaluate the vision encoder, we follow \cite{Gatys2015ANA, 10.1145/3512353.3512366} which perform Neural Style Transfer by matching the inner product of CNN filter responses. Similarly, we extract art style information by computing this inner product to obtain the Gram matrix of last-layer ViT patch features. The resulting features capture spatially-decorrelated texture information representing the art style of an image. The vision model then predicts by choosing the image option with the lowest MSE between these style features. We evaluate the VLM through a multiple-choice VQA format as specified by the BLINK benchmark.

\vspace{-2mm}
\section{Initial Observations}
\label{sec:results}

\begin{table*}[]
\scalebox{0.525}{
\begin{tabular}{@{}rccc|ccc|ccc|ccc|ccc|ccc@{}}
\toprule
\multicolumn{1}{c}{\textbf{}} & \multicolumn{3}{c|}{\textbf{Semantic Correspondence}} & \multicolumn{3}{c|}{\textbf{Low-level Matching}} & \multicolumn{3}{c|}{\textbf{Depth Estimation}} & \multicolumn{3}{c|}{\textbf{Object Affordances}} & \multicolumn{3}{c|}{\textbf{Art Style}} & \multicolumn{3}{c}{\textbf{3D Object Awareness}} \\ \midrule
\multicolumn{1}{c}{\textbf{}} & \textbf{Visual} & \textbf{VLM} & \textbf{\begin{tabular}[c]{@{}c@{}}VLM\\ blind\end{tabular}} & \textbf{Visual} & \textbf{VLM} & \textbf{\begin{tabular}[c]{@{}c@{}}VLM\\ blind\end{tabular}} & \textbf{Visual} & \textbf{VLM} & \textbf{\begin{tabular}[c]{@{}c@{}}VLM\\ blind\end{tabular}} & \textbf{Visual} & \textbf{VLM} & \textbf{\begin{tabular}[c]{@{}c@{}}VLM\\ blind\end{tabular}} & \textbf{Visual} & \textbf{VLM} & \textbf{\begin{tabular}[c]{@{}c@{}}VLM\\ blind\end{tabular}} & \textbf{Visual} & \multicolumn{1}{l}{\textbf{VLM}} & \multicolumn{1}{l}{\textbf{\begin{tabular}[c]{@{}c@{}}VLM\\ blind\end{tabular}}} \\ \midrule
\textbf{DINOv2} & \boldone{0.536} & 0.225 & 0.298 & \boldone{0.819} & 0.170 & 0.251 & \boldone{0.887} & 0.653 & 0.597 & \boldone{0.411} & 0.155 & 0.154 & 0.675 & 0.530 & \boldthree{0.446} & \boldone{0.598} & 0.315 & 0.311 \\
\textbf{IN-1k} & 0.428 & \boldtwo{0.304} & \boldthree{0.316} & 0.719 & \boldtwo{0.287} & 0.243 & 0.870 & 0.640 & 0.593 & 0.380 & \boldtwo{0.178} & \boldthree{0.168} & 0.744 & 0.530 & 0.440 & 0.482 & 0.311 & \boldthree{0.313} \\
\textbf{CLIP} & 0.384 & 0.297 & 0.298 & 0.532 & 0.181 & \boldthree{0.245} & 0.852 & 0.623 & \boldthree{0.618} & 0.295 & 0.171 & 0.149 & \boldone{0.761} & \boldtwo{0.538} & 0.439 & 0.503 & \boldtwo{0.328} & 0.307 \\
\textbf{SigLIP} & 0.391 & \boldtwo{0.304} & 0.315 & 0.632 & 0.246 & 0.240 & 0.840 & \boldtwo{0.663} & 0.612 & 0.248 & 0.147 & 0.160 & 0.744 & 0.427 & 0.428 & 0.587 & 0.320 & 0.307 \\ \bottomrule
\end{tabular}
}
\caption{\small \textbf{Results of visual evaluation, VLM evaluation, and blind evaluation.} Across all tasks the visual evaluation approach performs better than both the VLM and blind evaluations by a wide margin, despite substantial variation in vision encoder performance. Furthermore, we see that despite \textbf{DINOv2} being the highest performing encoder in 5 of 6 tasks it does not lead to the highest performing VLM approach in any task, highlighting a rank order shift from standalone visual encoders to VLMs. 
\vspace{-3mm}
}
\end{table*}
\label{tab:full_results}

\vspace{-2mm}
\subsection{Large-scale vision encoders perform well}

\vspace{-1mm}
 We report results in Fig. \ref{fig:main_results} and Tab. \ref{tab:full_results}; large vision models perform well above chance on all tasks (e.g., $88.7\%$ on depth estimation). This performance is unsurprising, as prior work has demonstrated that large vision encoders (e.g., DINOv2) excel at vision-centric tasks. We also observe significant variation in performance between visual encoders, with DINOv2 consistently outperforming others on all VQA tasks requiring an element of spatial awareness (all tasks excluding `Art Style'). Our results corroborate prior work which demonstrate that vision encoders extract representations useful on a variety of vision-centric tasks \cite{bananiProbing3DAwareness2024, zhanGeneralProtocolProbe2024}. Moreover, the rank ordering of model performance here is consistent with prior results; vision-only pretraining generates models that often outperform those with vision-langauge pretraining on spatially-dependent tasks \cite{bananiProbing3DAwareness2024}. We emphasize that, as our evaluation approaches are zero-shot except for depth estimation where a DPT head is trained, the representations required for these tasks are readily-accessible in these visual encoders. As such, task-relevant visual information should also be easy to access for downstream models using these visual representations. 

\vspace{-2mm}
\subsection{VLMs perform far worse than their encoders}
\vspace{-1mm}

Our primary observation is the universal drop in performance when moving from a standard visual evaluation strategy to a VLM: in many cases, performance degrades from near-ceiling to chance levels as evidenced in Figure \ref{fig:main_results} and Table \ref{tab:full_results}.
The low-level matching task suffers the steepest drop at $45.5\%$, and depth estimation drops the least ($21.7\%$).
These results indicate that something - likely either the quality of visual representations or the ability to use them - is lost between the vision encoder and the LLM output, as information easily-extractable from vision representations seems to `disappear' before the VQA task is completed.

Throughout this work, we focus our evaluations on a suite of VLMs that do not tune the vision encoder, allowing us to vary the way that a fixed set of vision representations are being used. 
However, we corroborate our results by also evaluating on commonly-used VLMs: InternVL \cite{chen2024internvl}, Phi-3-V \cite{abdin2024phi3technicalreporthighly}, and QwenVL \cite{bai2023qwenvlversatilevisionlanguagemodel}. As seen in Figure \ref{fig:sota_results}, we observe the same type of performance drop when switching from probing the VLM's vision encoder to a VQA task. We note one exception with InternVL: when the vision representations themselves cannot easily complete the task, the VLM cannot rescue performance.
These results are especially compelling when considering that these vision encoders were trained to be used within the corresponding VLM, not as a standalone.

\vspace{-2mm}
\subsection{VLMs change rank ordering of vision encoders}
\vspace{-1mm}

Our next observation comes from the change in the ranking of model performance, when comparing a vision-only evaluation to VLM performance. In almost every task DINOv2 performs best when using a direct visual readout. 
This result is aligned with previous benchmarking results across multiple vision backbones \cite{bananiProbing3DAwareness2024, zhanGeneralProtocolProbe2024}, which find that DINOv2 excels at these tasks. However, when transitioning to a VLM evaluation using the same visual encoders, DINOv2 experiences a disproportionate decrease in performance for several tasks. This asymmetric drop leads to a different ranking of which visual encoders enable the most capable VLMs. These VLM-based rankings have already led to conclusions that CLIP-style pretraining strategies are the most promising for VLMs \cite{tongCambrian1FullyOpen2024}. Our results suggest this might be a premature conclusion.

\vspace{-2mm}
\subsection{VLM choices reflect their `blind' baselines}
\begin{figure}[h]
    \centering
\vspace{-2mm}
    \includegraphics[width=1\linewidth]{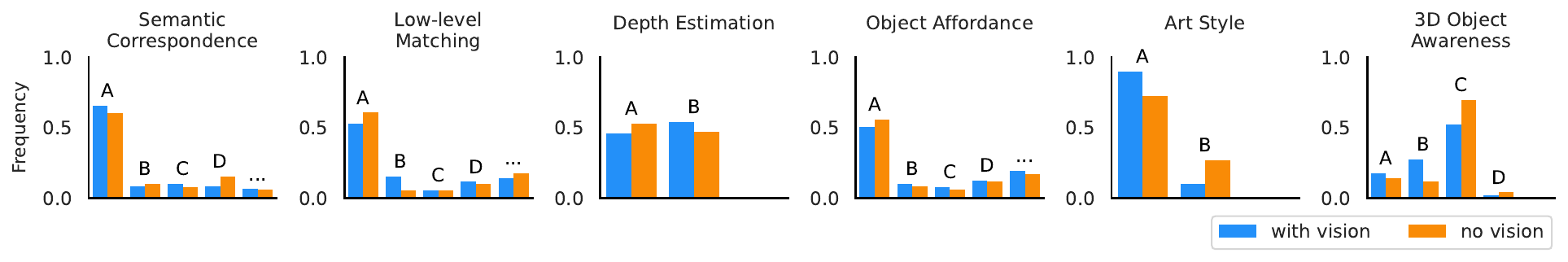}
\vspace{-7mm}
    \caption
    {
\small \textbf{VLM choice behavior reflects the biases of their LLMs.} Here we visualize the distribution of answers when models are presented with (blue) and without (orange) a valid image.  We find that behaviors largely reflect the pattern of choices in the blind baselines. We take this as evidence that VLMs are not simply misuing their visual representations, but they inherit their blind biases. 
   \vspace{-2mm}
    }
    \label{fig:blind_mc}
\end{figure}

\vspace{-1mm}
\label{subsec:language_prior}
The chance-level performance of VLMs on these vision-centric tasks, alongside the near ceiling performance of their visual encoders, raises the possibility that VLMs are not relying on their visual inputs to generate choice behaviors. We explore this possibility first by visualizing the distribution of multiple-choice answer outputs. Interestingly, the choice behaviors of VLMs are non-uniform (Fig. \ref{fig:blind_mc}, `with vision'). We next repeat all experiments in each benchmark with a blank visual input to the VLM along with the prompt, and find that the VLMs output answer distributions are uncannily similar to their `blinded' counterparts (Fig. \ref{fig:blind_mc}, `no vision'). These results suggest that VLMs ignore images while inheriting the biases within their LLM components. We also provide quantitative measurements of these differences in distribution by measuring total variation (TV) distance in Table \ref{tab:tv_results}.
\vspace{-4mm}

\section{Analysis of VLM performance}
\vspace{-1mm}
\subsection{Vision features maintain their integrity throughout VLM layers}
\begin{wrapfigure}{r}{0.58\textwidth}
  \begin{center}
    \vspace{-10mm}
    \includegraphics[width=0.58\textwidth]{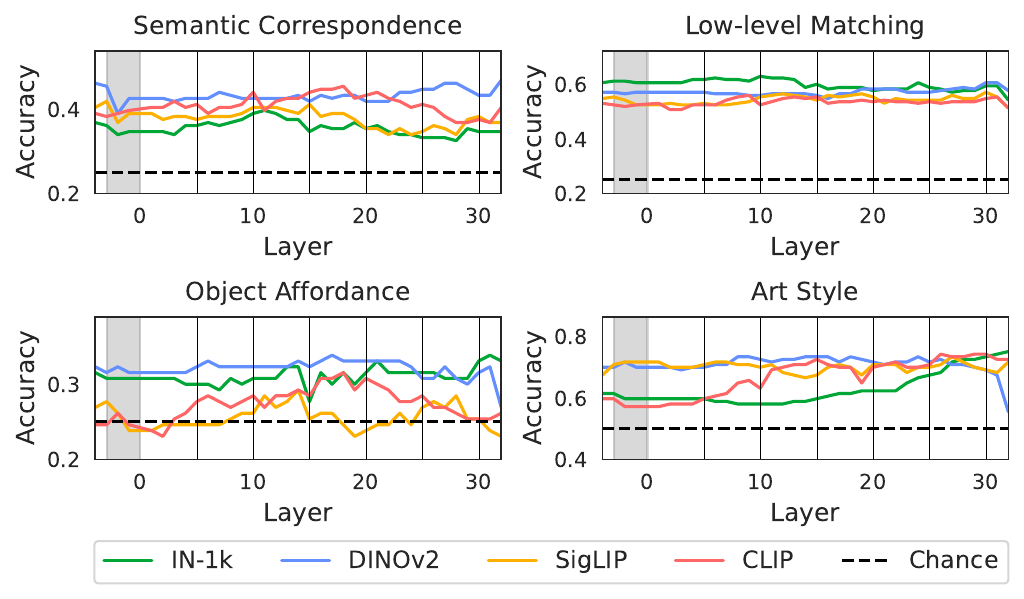}
    \vspace{-8mm}
  \end{center}

  \caption
    {\small \textbf{Visual evaluations for intermediate VLM layers. } We probe vision representations throughout the projector (gray region) and LLM (white region) layers, finding that they generally preserve task-relevant information and show no significant degradation.
    \vspace{-3mm}
    }
    \label{fig:layer_probe}
\end{wrapfigure}

\vspace{-1mm}
\label{subsec:vit}
One possible explanation for the poor VLM performance is the degradation of vision representations — specifically, that transformations applied by the projector and/or LLM layers may discard task-relevant visual information. 
If visual evaluation strategies remain effective throughout the rest of the VLM, it would indicate that the relevant information is preserved and accessible at every stage.
Note that this is a sufficient but not necessary condition; even if vision representations are not amenable to our probing methods, they may still be useful to the LLM through a different algorithm. Nonetheless, this condition is met; as shown in Fig. \ref{fig:layer_probe}, vision representations remain intact as they pass through the projector (gray region) and LLM (white region), suggesting that performance drops are not due to a loss of visual information.

We highlight two interesting trends/exceptions: first, the largest jumps in performance tend to happen in the final layer of the LLM. DINOv2's vision representations retain signal in the Object Affordance and Art Style tasks until the last layer, where they decline sharply. We also observe a similar but smaller drop in all final layers for Low-level Matching. We hypothesize that these shifts come from the LLM's shift in priority from preserving and attending to useful features to generating a natural language answer in the final layer. Second, the VLM using an ImageNet-supervised ViT (IN-1k) actually \textit{better} encodes art style in later layers of the LLM, yet its final performance is only 53\%. This observation further emphasizes the high quality of vision representations within the VLM and its mismatched VQA performance. 

\vspace{-1mm}
\subsection{Prompt-tuning results in minimal improvements}
\vspace{-1mm}
\label{subsec:prompt}

\begin{figure}[h]
  \begin{center}
    \vspace{-2mm}
    \includegraphics[width=1.0\textwidth]{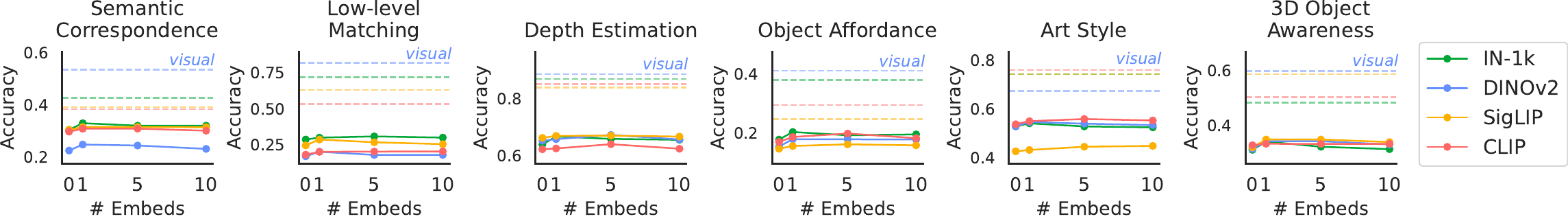}
    \vspace{-6mm}
  \end{center}

  \caption{
    \small \textbf{Prompt-tuning evaluation.} We tune [1, 5, 10] prefix embeddings and compare results with the original performance (\texttt{x=0}) and visual evaluation ceiling (dotted line). We observe minimal returns that diminish after 1-5 prefix embeddings. 
    \vspace{-2mm}
    }
    \label{fig:fewshot}
\end{figure}

If vision representations are not the primary culprit of low VLM performance, is it possible that the models are very sensitive to how they are prompted? 
We evaluate this possibility by introducing learnable prefixes passed along with the original prompt embeddings.
Following \cite{lesterPowerScaleParameterEfficient2021}, we prepend the original prompt embeddings with randomly-initialized trainable vectors and tune on 1,000 VQA examples of the task. Keeping the VLM and original prompt frozen, we minimize cross-entropy between the next-token distribution and ground truth. We report results in Fig. \ref{fig:fewshot}, and training details in the Supplement. 

We find that adding a single prefix embedding improves performance only marginally (Fig. \ref{fig:fewshot}), especially considering how VLM performance is hovering around chance. As shown in the same figure, tuning additional (5, 10) prompt embeddings does not continue improving performance. 
Our results indicate that VLMs still struggle to successfully perform vision-centric tasks despite having an encoding of the task at hand, eliminating sensitivity to input prompt as a performance bottleneck. 
\vspace{-2mm}
\subsection{The language model underutilizes its vision representations}
\label{subsec:lang_ft}
\begin{figure}[t]
    \centering
    \includegraphics[width=1\linewidth]{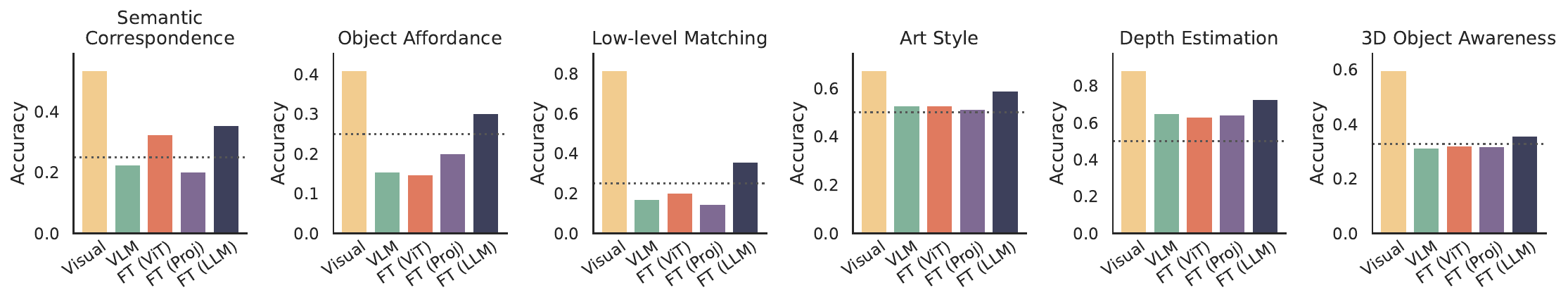}
    \vspace{-7mm}
    \caption
    {\small 
    We find that tuning the LLM (as opposed to the projector layers or the ViT) on each task, with the same parameter count in each setting, provides the largest performance increase. Taken together with Fig. \ref{fig:attn}, these results provide evidence that the LLM's ability to use its visual representations is a bottleneck in succeeding on vision-centric tasks.
    \vspace{-3mm}
    }
    \label{fig:ft}
\end{figure}

Having ruled out degradation of vision representations with the VLM (Sec. \ref{subsec:vit}) and the sensitivity to prompt formulation (Sec. \ref{subsec:prompt}) as performance bottlenecks, we now examine the role of the LLM itself. We finetune the individual components of the VLM - the ViT, projector, and LLM - on 5,000 examples of each vision-centric task. Using the same VQA format as our evaluation suite, we apply LoRA-tuning while controlling tunable weight matrices to ensure equal parameter counts across components (16.7M parameters, equivalent to full projector fine-tuning). See the Supplement for further training and implementation details.

\begin{wrapfigure}{r}{0.51\textwidth}
  \begin{center}
    \vspace{-6mm}
    \includegraphics[width=0.51\textwidth]{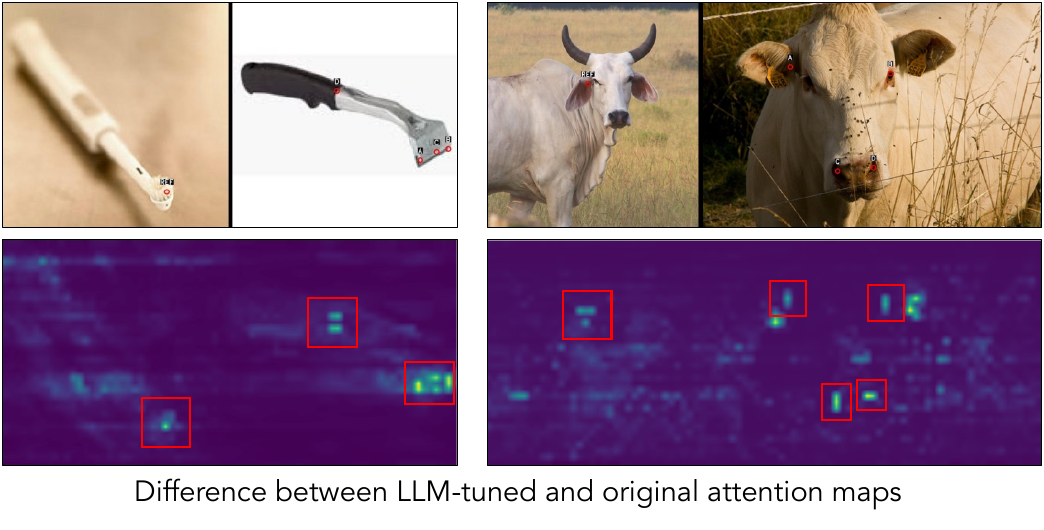}
    \vspace{-8mm}
  \end{center}

  \caption{\small  Visualizing the difference between attention maps before and after fine-tuning the LLM, we observe an increase in attention at the points of interest (\texttt{REF}, \texttt{A}, \texttt{B}, \texttt{C}, \texttt{D}) for correspondence tasks. These points are most salient in attention layers 4-6; here we visualize layer 4 for Object Affordance (left) and Semantic Correspondence (right).
  \vspace{-2mm}
  }
  \label{fig:attn}
\end{wrapfigure}

As shown in Fig. \ref{fig:ft}, tuning the LLM provides the largest gains in performance over tuning the projector or ViT layers. To better understand this effect, we analyze attention shifts over DINOv2 representations after fine-tuning. 
In correspondence tasks, fine-tuning enhances attention over multiple-choice labels, reference points, and other text (particularly in LLM layer 4). Unlike projector or ViT tuning, which does not consistently highlight these points in any attention layer or head, LLM tuning improves the model's ability to find and use vision representations at key regions.
We also note an exception in the 3D Object Awareness task, where the performance improvements are less notable. We hypothesize two reasons: first, unlike other tasks requiring localized feature extraction (e.g., bounding boxes, texture), 3D Object Awareness demands more abstract, variable features that lightweight fine-tuning may not fully capture. Second, our ShapeNet-based data is not a perfect reflection of the MOCHI benchmark, which includes ShapeNet renderings as well as a broader range of objects and settings. 
More attention visualizations are provided in the Supplement to further illustrate these effects.
\vspace{-1mm}
\subsubsection{Tuning the LLM alleviates the language prior}
\label{subsec:langprior_ft}
We observed in Section \ref{subsec:language_prior} that not only do VLMs fail on vision-centric tasks, they inherit their language biases. Given the previous results that tuning the LLM best improves performance, we revisit this observation by evaluating TV distance between prediction and ground truth distributions. As seen in Table \ref{tab:tv_results}, the LLM-tuned models produce answers with the lowest TV distance to the ground truth. The only exception is on Depth Estimation, where the original VLM already closely matched the ground truth distribution.

Taking the results from Sections \ref{subsec:lang_ft} and \ref{subsec:langprior_ft}, we find that the language model is the limiting factor in making accurate predictions for vision-centric tasks, and that its limitations come from both a lack of attention on the salient visual regions and a heavy reliance on the biases toward certain multiple choice answers present in the LLM.
Note that we do not propose training directly on the tasks as an overarching solution to the lack of VLMs' abilities to use their visual representations; rather, we use this method to help localize failures in the VLM and how improving the LLM's ability to use its vision representations can overcome its language prior and most effectively boost performance.

\begin{table}[h]
\centering
\scalebox{0.75}{
\begin{tabular}{@{}rcccccc@{}}
\toprule
\multicolumn{1}{l}{\textbf{}} & \textbf{\begin{tabular}[c]{@{}c@{}}Semantic\\ Correspondence\end{tabular}} & \textbf{\begin{tabular}[c]{@{}c@{}}Low-level\\ Matching\end{tabular}} & \textbf{\begin{tabular}[c]{@{}c@{}}Depth\\ Estimation\end{tabular}} & \textbf{\begin{tabular}[c]{@{}c@{}}Object\\ Affordance\end{tabular}} & \textbf{\begin{tabular}[c]{@{}c@{}}Art\\ Style\end{tabular}} & \textbf{\begin{tabular}[c]{@{}c@{}}3D Object\\ Awareness\end{tabular}} \\ \midrule
\textbf{Original} & 0.337 & 0.494 & 0.043 & 0.521 & 0.259 & 0.386 \\
\textbf{Blind} & 0.399 & 0.380 & 0.101 & 0.488 & 0.427 & 0.195 \\
\textbf{FT (ViT)} & 0.393 & 0.500 & 0.146 & 0.593 & 0.474 & 0.161 \\
\textbf{FT (projector)} & 0.391 & 0.364 & \textbf{0.040} & 0.391 & 0.118 & 0.105 \\
\textbf{FT (LLM)} & \textbf{0.078} & \textbf{0.121} & 0.103 & \textbf{0.190} & \textbf{0.071} & \textbf{0.069} \\ \bottomrule
\end{tabular}
}
\caption{\small \textbf{Total variation distance between VLM predictions and GT answers. } Fine-tuning the LLM on each task brings the distribution of multiple choice answers closer to the ground truth distribution, lessening the effect of the language prior. Meanwhile, fine-tuning either the ViT or the projector layers do little to improve the blind bias toward certain multiple-choice answers.}
\end{table}
\label{tab:tv_results}

\vspace{-4mm}
\section{Related Work}
\label{sec:related}
\vspace{-2mm}
Visual question-answering (VQA) tasks are commonly used to evaluate VLM performance in response to an image and a text prompt. Of particular interest here are `vision-centric' benchmarks \cite{tongCambrian1FullyOpen2024, fuBLINKMultimodalLarge2024}, which often reformat computer vision tasks into VQA benchmarks. For example, a multi-view correspondence task (Fig. \ref{fig:teaser}) can be converted into a multiple choice format by presenting a pair of images and asking questions such as "Which point in image two best corresponds to the reference point in image one---A, B, C, or D?" %
This approach has been used to convert many computer vision tasks, including depth estimation, object affordances, semantic and visual similarity, etc. 

\vspace{-1mm}
\subsection{Vision-language models}
\vspace{-1mm}
Late-fusion VLM architectures have largely converged on three components: a vision backbone, a language model processing image and prompt tokens, and an adapter aligning the vision representations to language space. Architectures and training strategies vary widely within this framework across model families such as LLaVA \cite{liuImprovedBaselinesVisual2023}, IDEFICS \cite{laurenconWhatMattersWhen2024}, PaliGemma \cite{beyer2024paligemmaversatile3bvlm}, Flamingo \cite{alayracFlamingoVisualLanguage2022}, and more \cite{shi2024eagleexploringdesignspace, fan2024mousipolyvisualexpertvisionlanguagemodels, awadalla2023openflamingoopensourceframeworktraining}; differences in backbone choice, adapter design, and instruction-tuning all contribute to vastly different performance patterns. Nevertheless, these models all rely on the existing representational power of their vision and language backbones. Most contemporary VLMs choose a vision-language pretrained vision encoder such as CLIP \cite{radfordLearningTransferableVisual2021} or SigLIP \cite{zhaiSigmoidLossLanguage2023}, as they have been shown to perform well \cite{karamchetiPrismaticVLMsInvestigating2024} and it is intuitive that a model previously aligned to some language space would be well-equipped to serve in multimodal contexts. Some works find ensembling vision backbones (or models specialized for certain tasks) to work best, as they complement each other and provide more expressive starting visual representations \cite{tong2024eyes, shi2024eagleexploringdesignspace, fan2024mousipolyvisualexpertvisionlanguagemodels}.

\vspace{-1mm}
\subsection{Studying VLM components}
\vspace{-1mm}
Recent works have questioned these design decisions and carefully studied the axes along which the VLMs vary. \cite{tongCambrian1FullyOpen2024, karamchetiPrismaticVLMsInvestigating2024, laurenconWhatMattersWhen2024} make several key findings about how VLM performance varies as a function of backbone and adapter choice, instruction tuning, image resolution, and other factors. Importantly, these works also study how VLM performance correlates with the strength of the vision and language backbones but do not converge on a clear answer; while \cite{karamchetiPrismaticVLMsInvestigating2024} finds that VLMs built on top of more performant LLMs do not necessarily score higher on VLM benchmarks, \cite{laurenconWhatMattersWhen2024} identifies the LLM as more of a bottleneck than the vision backbone. 

Each of these works analyzing VLM design axes distill their findings into a new family of multimodal models that improve over existing ones on VQA-style benchmarks. Many of these benchmarks emphasize the LLM's existing knowledge and reasoning abilities in tandem with the vision representations, yet interpret these results to reflect vision ability in the VLM \cite{tongCambrian1FullyOpen2024, karamchetiPrismaticVLMsInvestigating2024}. We further explore how indicative these benchmarks are of VLMs' vision representations, focusing on `vision-centric' tasks that evaluate visual perception abilities and that can be decoupled from language.

\subsubsection{Vision-centric VLM benchmarks. } \cite{fuBLINKMultimodalLarge2024} introduces the BLINK benchmark, a collection of visual perception tasks that can be solved by humans `within a blink' but are still challenging for VLMs. BLINK casts more traditional visual tasks (e.g., semantic correspondence, art style) in a multiple-choice format to probe visual abilities through language and finds that VLMs still struggle at these tasks. Similarly, \cite{tongCambrian1FullyOpen2024} presents CV-Bench as a re-casting of standard vision tasks (spatial relationship, object counting, depth order, distance) into a multimodal context. Evaluations on CV-Bench reveal language-supervised models (e.g., CLIP, SigLIP) as more performant vision backbones than self-supervised models (e.g., DINOv2, MoCov3 \cite{chen2021empiricalstudytrainingselfsupervised}); these findings contrast with recent benchmarking with visual evaluation strategies \cite{bananiProbing3DAwareness2024, zhanGeneralProtocolProbe2024} which find the inverse relationship. Our paper investigates the inconsistencies foreshadowed in these works, and further explores how indicative these VLM benchmarks truly are of vision model performance.

\section{Discussion}
\label{sec:discussion}

Despite their impressive performance on knowledge-based benchmarks, VLMs remain in some respects unaware of their own `sight'. 
Our work highlights a critical disconnect between the representations from vision encoders and their underuse within VLMs, leading to markedly suboptimal performance on vision-centric tasks. %
While vision representations perform well when evaluated independently, their integration into VLMs severely degrades vision-centric capabilities and causes them to rely on their language priors. 

Previous works have 1) attributed VLM limitations to vision encoder weakness and proposed ensembling encoders to mitigate these shortcomings \cite{karamchetiPrismaticVLMsInvestigating2024, tong2024eyes}. 
Our work suggests that this strategy is unlikely to address the underlying failures of VLMs to use their visual representations, as the vision encoder is not the bottleneck.
\cite{tongCambrian1FullyOpen2024} has also offered VLMs as an interface with which to evaluate visual representations. While this strategy allows for a much broader range of benchmarking tasks, we caution against drawing conclusions about vision representations when their VLM-based evaluations produce 1) near-chance performance and 2) changes in rank-ordering compared to a direct visual readout. 
While previous works have taken VLM results to suggest that certain vision encoders consistently offer advantages over other types of visual encoders (often, CLIP/SigLIP $>>$ DINOv2), our results show a lack of correlation between vision encoder rankings when using visual and VLM evaluation strategies.
We suggest that the perceived limitations of VLMs' vision encoders may simply come from their misuse rather than an inherent inability of the vision representations to perform the task at hand.

While our work has identified how poor performance in VLMs is a consequence of their inability to integrate visual information into its linguistic components, that does not necessarily imply that current visual encoders are sufficient for all visual tasks. In fact, there are well-documented shortcomings of contemporary large-scale vision encoders. For example, humans consistently outperform vision models on 3D shape inferences \cite{bonnen2024evaluatingmultiviewobjectconsistency}, indicating there there are still improvements needed. While we have argued above that the failure modes of VLMs are better attributed to the visual-linguistic integration rather than the failure of the visual encoder itself, we stand with prior work highlighting the importance of developing stronger vision encoders, as their abilities work in tandem with a capable LLM to succeed on vision-centric VQA tasks.

Finally, while we present a critical assessment of current VLMs, language remains a powerful interface to specify and evaluate performance. 
Indeed, many tasks are difficult to specify through an image alone, and language might allow us to probe models on a more diverse and complex set of visual abilities.
Yet, our results call for caution when making VLM design choices or drawing conclusions about their visual abilities through language benchmarks, as they may underestimate and misrepresent the actual capabilities of vision encoders.

\bibliography{colm2025_conference}

\begin{thebibliography}{38}
\providecommand{\natexlab}[1]{#1}
\providecommand{\url}[1]{\texttt{#1}}
\expandafter\ifx\csname urlstyle\endcsname\relax
  \providecommand{\doi}[1]{doi: #1}\else
  \providecommand{\doi}{doi: \begingroup \urlstyle{rm}\Url}\fi

\bibitem[Abdin et~al.(2024)Abdin, Aneja, Awadalla, Awadallah, Awan, Bach, Bahree, Bakhtiari, Bao, Behl, Benhaim, Bilenko, Bjorck, Bubeck, Cai, Cai, Chaudhary, Chen, Chen, Chen, Chen, Chen, Cheng, Chopra, Dai, Dixon, Eldan, Fragoso, Gao, Gao, Gao, Garg, Giorno, Goswami, Gunasekar, Haider, Hao, Hewett, Hu, Huynh, Iter, Jacobs, Javaheripi, Jin, Karampatziakis, Kauffmann, Khademi, Kim, Kim, Kurilenko, Lee, Lee, Li, Li, Liang, Liden, Lin, Lin, Liu, Liu, Liu, Liu, Liu, Luo, Madan, Mahmoudzadeh, Majercak, Mazzola, Mendes, Mitra, Modi, Nguyen, Norick, Patra, Perez-Becker, Portet, Pryzant, Qin, Radmilac, Ren, de~Rosa, Rosset, Roy, Ruwase, Saarikivi, Saied, Salim, Santacroce, Shah, Shang, Sharma, Shen, Shukla, Song, Tanaka, Tupini, Vaddamanu, Wang, Wang, Wang, Wang, Wang, Wang, Ward, Wen, Witte, Wu, Wu, Wyatt, Xiao, Xu, Xu, Xu, Xue, Yadav, Yang, Yang, Yang, Yang, Yu, Yuan, Zhang, Zhang, Zhang, Zhang, Zhang, Zhang, Zhang, and Zhou]{abdin2024phi3technicalreporthighly}
Marah Abdin, Jyoti Aneja, Hany Awadalla, Ahmed Awadallah, Ammar~Ahmad Awan, Nguyen Bach, Amit Bahree, Arash Bakhtiari, Jianmin Bao, Harkirat Behl, Alon Benhaim, Misha Bilenko, Johan Bjorck, Sébastien Bubeck, Martin Cai, Qin Cai, Vishrav Chaudhary, Dong Chen, Dongdong Chen, Weizhu Chen, Yen-Chun Chen, Yi-Ling Chen, Hao Cheng, Parul Chopra, Xiyang Dai, Matthew Dixon, Ronen Eldan, Victor Fragoso, Jianfeng Gao, Mei Gao, Min Gao, Amit Garg, Allie~Del Giorno, Abhishek Goswami, Suriya Gunasekar, Emman Haider, Junheng Hao, Russell~J. Hewett, Wenxiang Hu, Jamie Huynh, Dan Iter, Sam~Ade Jacobs, Mojan Javaheripi, Xin Jin, Nikos Karampatziakis, Piero Kauffmann, Mahoud Khademi, Dongwoo Kim, Young~Jin Kim, Lev Kurilenko, James~R. Lee, Yin~Tat Lee, Yuanzhi Li, Yunsheng Li, Chen Liang, Lars Liden, Xihui Lin, Zeqi Lin, Ce~Liu, Liyuan Liu, Mengchen Liu, Weishung Liu, Xiaodong Liu, Chong Luo, Piyush Madan, Ali Mahmoudzadeh, David Majercak, Matt Mazzola, Caio César~Teodoro Mendes, Arindam Mitra, Hardik Modi, Anh Nguyen,
  Brandon Norick, Barun Patra, Daniel Perez-Becker, Thomas Portet, Reid Pryzant, Heyang Qin, Marko Radmilac, Liliang Ren, Gustavo de~Rosa, Corby Rosset, Sambudha Roy, Olatunji Ruwase, Olli Saarikivi, Amin Saied, Adil Salim, Michael Santacroce, Shital Shah, Ning Shang, Hiteshi Sharma, Yelong Shen, Swadheen Shukla, Xia Song, Masahiro Tanaka, Andrea Tupini, Praneetha Vaddamanu, Chunyu Wang, Guanhua Wang, Lijuan Wang, Shuohang Wang, Xin Wang, Yu~Wang, Rachel Ward, Wen Wen, Philipp Witte, Haiping Wu, Xiaoxia Wu, Michael Wyatt, Bin Xiao, Can Xu, Jiahang Xu, Weijian Xu, Jilong Xue, Sonali Yadav, Fan Yang, Jianwei Yang, Yifan Yang, Ziyi Yang, Donghan Yu, Lu~Yuan, Chenruidong Zhang, Cyril Zhang, Jianwen Zhang, Li~Lyna Zhang, Yi~Zhang, Yue Zhang, Yunan Zhang, and Xiren Zhou.
\newblock Phi-3 technical report: A highly capable language model locally on your phone, 2024.
\newblock URL \url{https://arxiv.org/abs/2404.14219}.

\bibitem[Alayrac et~al.(2022)Alayrac, Donahue, Luc, Miech, Barr, Hasson, Lenc, Mensch, Millican, Reynolds, Ring, Rutherford, Cabi, Han, Gong, Samangooei, Monteiro, Menick, Borgeaud, Brock, Nematzadeh, Sharifzadeh, Binkowski, Barreira, Vinyals, Zisserman, and Simonyan]{alayracFlamingoVisualLanguage2022}
Jean-Baptiste Alayrac, Jeff Donahue, Pauline Luc, Antoine Miech, Iain Barr, Yana Hasson, Karel Lenc, Arthur Mensch, Katie Millican, Malcolm Reynolds, Roman Ring, Eliza Rutherford, Serkan Cabi, Tengda Han, Zhitao Gong, Sina Samangooei, Marianne Monteiro, Jacob Menick, Sebastian Borgeaud, Andrew Brock, Aida Nematzadeh, Sahand Sharifzadeh, Mikolaj Binkowski, Ricardo Barreira, Oriol Vinyals, Andrew Zisserman, and Karen Simonyan.
\newblock Flamingo: a {Visual} {Language} {Model} for {Few}-{Shot} {Learning}, November 2022.
\newblock URL \url{http://arxiv.org/abs/2204.14198}.
\newblock arXiv:2204.14198 [cs].

\bibitem[Awadalla et~al.(2023)Awadalla, Gao, Gardner, Hessel, Hanafy, Zhu, Marathe, Bitton, Gadre, Sagawa, Jitsev, Kornblith, Koh, Ilharco, Wortsman, and Schmidt]{awadalla2023openflamingoopensourceframeworktraining}
Anas Awadalla, Irena Gao, Josh Gardner, Jack Hessel, Yusuf Hanafy, Wanrong Zhu, Kalyani Marathe, Yonatan Bitton, Samir Gadre, Shiori Sagawa, Jenia Jitsev, Simon Kornblith, Pang~Wei Koh, Gabriel Ilharco, Mitchell Wortsman, and Ludwig Schmidt.
\newblock Openflamingo: An open-source framework for training large autoregressive vision-language models, 2023.
\newblock URL \url{https://arxiv.org/abs/2308.01390}.

\bibitem[Bai et~al.(2023)Bai, Bai, Yang, Wang, Tan, Wang, Lin, Zhou, and Zhou]{bai2023qwenvlversatilevisionlanguagemodel}
Jinze Bai, Shuai Bai, Shusheng Yang, Shijie Wang, Sinan Tan, Peng Wang, Junyang Lin, Chang Zhou, and Jingren Zhou.
\newblock Qwen-vl: A versatile vision-language model for understanding, localization, text reading, and beyond, 2023.
\newblock URL \url{https://arxiv.org/abs/2308.12966}.

\bibitem[Balntas et~al.(2017)Balntas, Lenc, Vedaldi, and Mikolajczyk]{hpatches_2017_cvpr}
Vassileios Balntas, Karel Lenc, Andrea Vedaldi, and Krystian Mikolajczyk.
\newblock Hpatches: A benchmark and evaluation of handcrafted and learned local descriptors.
\newblock In \emph{CVPR}, 2017.

\bibitem[Banani et~al.(2024)Banani, Raj, Maninis, Kar, Li, Rubinstein, Sun, Guibas, Johnson, and Jampani]{bananiProbing3DAwareness2024}
Mohamed~El Banani, Amit Raj, Kevis-Kokitsi Maninis, Abhishek Kar, Yuanzhen Li, Michael Rubinstein, Deqing Sun, Leonidas Guibas, Justin Johnson, and Varun Jampani.
\newblock Probing the {3D} {Awareness} of {Visual} {Foundation} {Models}, April 2024.
\newblock URL \url{http://arxiv.org/abs/2404.08636}.
\newblock arXiv:2404.08636 [cs].

\bibitem[Beyer et~al.(2024)Beyer, Steiner, Pinto, Kolesnikov, Wang, Salz, Neumann, Alabdulmohsin, Tschannen, Bugliarello, Unterthiner, Keysers, Koppula, Liu, Grycner, Gritsenko, Houlsby, Kumar, Rong, Eisenschlos, Kabra, Bauer, Bošnjak, Chen, Minderer, Voigtlaender, Bica, Balazevic, Puigcerver, Papalampidi, Henaff, Xiong, Soricut, Harmsen, and Zhai]{beyer2024paligemmaversatile3bvlm}
Lucas Beyer, Andreas Steiner, André~Susano Pinto, Alexander Kolesnikov, Xiao Wang, Daniel Salz, Maxim Neumann, Ibrahim Alabdulmohsin, Michael Tschannen, Emanuele Bugliarello, Thomas Unterthiner, Daniel Keysers, Skanda Koppula, Fangyu Liu, Adam Grycner, Alexey Gritsenko, Neil Houlsby, Manoj Kumar, Keran Rong, Julian Eisenschlos, Rishabh Kabra, Matthias Bauer, Matko Bošnjak, Xi~Chen, Matthias Minderer, Paul Voigtlaender, Ioana Bica, Ivana Balazevic, Joan Puigcerver, Pinelopi Papalampidi, Olivier Henaff, Xi~Xiong, Radu Soricut, Jeremiah Harmsen, and Xiaohua Zhai.
\newblock Paligemma: A versatile 3b vlm for transfer, 2024.
\newblock URL \url{https://arxiv.org/abs/2407.07726}.

\bibitem[Bonnen et~al.(2024)Bonnen, Fu, Bai, O'Connell, Friedman, Kanwisher, Tenenbaum, and Efros]{bonnen2024evaluatingmultiviewobjectconsistency}
Tyler Bonnen, Stephanie Fu, Yutong Bai, Thomas O'Connell, Yoni Friedman, Nancy Kanwisher, Joshua~B. Tenenbaum, and Alexei~A. Efros.
\newblock Evaluating multiview object consistency in humans and image models, 2024.
\newblock URL \url{https://arxiv.org/abs/2409.05862}.

\bibitem[Brazil et~al.(2023)Brazil, Kumar, Straub, Ravi, Johnson, and Gkioxari]{brazil2023omni3dlargebenchmarkmodel}
Garrick Brazil, Abhinav Kumar, Julian Straub, Nikhila Ravi, Justin Johnson, and Georgia Gkioxari.
\newblock Omni3d: A large benchmark and model for 3d object detection in the wild, 2023.
\newblock URL \url{https://arxiv.org/abs/2207.10660}.

\bibitem[Chang et~al.(2015)Chang, Funkhouser, Guibas, Hanrahan, Huang, Li, Savarese, Savva, Song, Su, Xiao, Yi, and Yu]{chang2015shapenetinformationrich3dmodel}
Angel~X. Chang, Thomas Funkhouser, Leonidas Guibas, Pat Hanrahan, Qixing Huang, Zimo Li, Silvio Savarese, Manolis Savva, Shuran Song, Hao Su, Jianxiong Xiao, Li~Yi, and Fisher Yu.
\newblock Shapenet: An information-rich 3d model repository, 2015.
\newblock URL \url{https://arxiv.org/abs/1512.03012}.

\bibitem[Chen et~al.(2021)Chen, Xie, and He]{chen2021empiricalstudytrainingselfsupervised}
Xinlei Chen, Saining Xie, and Kaiming He.
\newblock An empirical study of training self-supervised vision transformers, 2021.
\newblock URL \url{https://arxiv.org/abs/2104.02057}.

\bibitem[Chen et~al.(2024)Chen, Wu, Wang, Su, Chen, Xing, Zhong, Zhang, Zhu, Lu, et~al.]{chen2024internvl}
Zhe Chen, Jiannan Wu, Wenhai Wang, Weijie Su, Guo Chen, Sen Xing, Muyan Zhong, Qinglong Zhang, Xizhou Zhu, Lewei Lu, et~al.
\newblock Internvl: Scaling up vision foundation models and aligning for generic visual-linguistic tasks.
\newblock In \emph{Proceedings of the IEEE/CVF Conference on Computer Vision and Pattern Recognition}, pp.\  24185--24198, 2024.

\bibitem[Chiang et~al.(2023)Chiang, Li, Lin, Sheng, Wu, Zhang, Zheng, Zhuang, Zhuang, Gonzalez, Stoica, and Xing]{vicuna2023}
Wei-Lin Chiang, Zhuohan Li, Zi~Lin, Ying Sheng, Zhanghao Wu, Hao Zhang, Lianmin Zheng, Siyuan Zhuang, Yonghao Zhuang, Joseph~E. Gonzalez, Ion Stoica, and Eric~P. Xing.
\newblock Vicuna: An open-source chatbot impressing gpt-4 with 90\%* chatgpt quality, March 2023.
\newblock URL \url{https://lmsys.org/blog/2023-03-30-vicuna/}.

\bibitem[Deng et~al.(2009)Deng, Dong, Socher, Li, Li, and Fei-Fei]{5206848}
Jia Deng, Wei Dong, Richard Socher, Li-Jia Li, Kai Li, and Li~Fei-Fei.
\newblock Imagenet: A large-scale hierarchical image database.
\newblock In \emph{2009 IEEE Conference on Computer Vision and Pattern Recognition}, pp.\  248--255, 2009.
\newblock \doi{10.1109/CVPR.2009.5206848}.

\bibitem[Dosovitskiy et~al.(2021)Dosovitskiy, Beyer, Kolesnikov, Weissenborn, Zhai, Unterthiner, Dehghani, Minderer, Heigold, Gelly, Uszkoreit, and Houlsby]{dosovitskiyImageWorth16x162021}
Alexey Dosovitskiy, Lucas Beyer, Alexander Kolesnikov, Dirk Weissenborn, Xiaohua Zhai, Thomas Unterthiner, Mostafa Dehghani, Matthias Minderer, Georg Heigold, Sylvain Gelly, Jakob Uszkoreit, and Neil Houlsby.
\newblock An {Image} is {Worth} 16x16 {Words}: {Transformers} for {Image} {Recognition} at {Scale}, June 2021.
\newblock URL \url{http://arxiv.org/abs/2010.11929}.
\newblock arXiv:2010.11929 [cs].

\bibitem[Fan et~al.(2024)Fan, Ji, Jiang, Li, Jin, Song, Wang, Hong, Chen, Zheng, Zhang, Huang, Zheng, Xi, Zhou, Dou, Ye, Yan, Gui, Zhang, Qiu, Huang, Wu, and Jiang]{fan2024mousipolyvisualexpertvisionlanguagemodels}
Xiaoran Fan, Tao Ji, Changhao Jiang, Shuo Li, Senjie Jin, Sirui Song, Junke Wang, Boyang Hong, Lu~Chen, Guodong Zheng, Ming Zhang, Caishuang Huang, Rui Zheng, Zhiheng Xi, Yuhao Zhou, Shihan Dou, Junjie Ye, Hang Yan, Tao Gui, Qi~Zhang, Xipeng Qiu, Xuanjing Huang, Zuxuan Wu, and Yu-Gang Jiang.
\newblock Mousi: Poly-visual-expert vision-language models, 2024.
\newblock URL \url{https://arxiv.org/abs/2401.17221}.

\bibitem[Fu et~al.(2024)Fu, Hu, Li, Feng, Wang, Lin, Roth, Smith, Ma, and Krishna]{fuBLINKMultimodalLarge2024}
Xingyu Fu, Yushi Hu, Bangzheng Li, Yu~Feng, Haoyu Wang, Xudong Lin, Dan Roth, Noah~A. Smith, Wei-Chiu Ma, and Ranjay Krishna.
\newblock {BLINK}: {Multimodal} {Large} {Language} {Models} {Can} {See} but {Not} {Perceive}, April 2024.
\newblock URL \url{http://arxiv.org/abs/2404.12390}.
\newblock arXiv:2404.12390 [cs].

\bibitem[Gatys et~al.(2015)Gatys, Ecker, and Bethge]{Gatys2015ANA}
Leon~A. Gatys, Alexander~S. Ecker, and Matthias Bethge.
\newblock A neural algorithm of artistic style.
\newblock \emph{ArXiv}, abs/1508.06576, 2015.
\newblock URL \url{https://api.semanticscholar.org/CorpusID:13914930}.

\bibitem[Karamcheti et~al.(2024)Karamcheti, Nair, Balakrishna, Liang, Kollar, and Sadigh]{karamchetiPrismaticVLMsInvestigating2024}
Siddharth Karamcheti, Suraj Nair, Ashwin Balakrishna, Percy Liang, Thomas Kollar, and Dorsa Sadigh.
\newblock Prismatic {VLMs}: {Investigating} the {Design} {Space} of {Visually}-{Conditioned} {Language} {Models}, May 2024.
\newblock URL \url{http://arxiv.org/abs/2402.07865}.
\newblock arXiv:2402.07865 [cs].

\bibitem[Lai et~al.(2021)Lai, Purushwalkam, and Gupta]{lai2021functional}
Zihang Lai, Senthil Purushwalkam, and Abhinav Gupta.
\newblock The functional correspondence problem.
\newblock In \emph{Proceedings of the IEEE/CVF International Conference on Computer Vision}, pp.\  15772--15781, 2021.

\bibitem[Laurençon et~al.(2024)Laurençon, Tronchon, Cord, and Sanh]{laurenconWhatMattersWhen2024}
Hugo Laurençon, Léo Tronchon, Matthieu Cord, and Victor Sanh.
\newblock What matters when building vision-language models?, May 2024.
\newblock URL \url{http://arxiv.org/abs/2405.02246}.
\newblock arXiv:2405.02246 [cs].

\bibitem[Lester et~al.(2021)Lester, Al-Rfou, and Constant]{lesterPowerScaleParameterEfficient2021}
Brian Lester, Rami Al-Rfou, and Noah Constant.
\newblock The {Power} of {Scale} for {Parameter}-{Efficient} {Prompt} {Tuning}, September 2021.
\newblock URL \url{http://arxiv.org/abs/2104.08691}.
\newblock arXiv:2104.08691 [cs].

\bibitem[Liu et~al.(2023)Liu, Li, Li, and Lee]{liuImprovedBaselinesVisual2023}
Haotian Liu, Chunyuan Li, Yuheng Li, and Yong~Jae Lee.
\newblock Improved {Baselines} with {Visual} {Instruction} {Tuning}, October 2023.
\newblock URL \url{http://arxiv.org/abs/2310.03744}.
\newblock arXiv:2310.03744 [cs].

\bibitem[Liu et~al.(2024)Liu, Li, Li, Li, Zhang, Shen, and Lee]{liu2024llavanext}
Haotian Liu, Chunyuan Li, Yuheng Li, Bo~Li, Yuanhan Zhang, Sheng Shen, and Yong~Jae Lee.
\newblock Llava-next: Improved reasoning, ocr, and world knowledge, January 2024.
\newblock URL \url{https://llava-vl.github.io/blog/2024-01-30-llava-next/}.

\bibitem[Lu(2022)]{10.1145/3512353.3512366}
Jiahao Lu.
\newblock Transformer-based neural texture synthesis and style transfer.
\newblock In \emph{Proceedings of the 2022 4th Asia Pacific Information Technology Conference}, APIT '22, pp.\  88–95, New York, NY, USA, 2022. Association for Computing Machinery.
\newblock ISBN 9781450395571.
\newblock \doi{10.1145/3512353.3512366}.
\newblock URL \url{https://doi.org/10.1145/3512353.3512366}.

\bibitem[Min et~al.(2019)Min, Lee, Ponce, and Cho]{min2019spair71klargescalebenchmarksemantic}
Juhong Min, Jongmin Lee, Jean Ponce, and Minsu Cho.
\newblock Spair-71k: A large-scale benchmark for semantic correspondence, 2019.
\newblock URL \url{https://arxiv.org/abs/1908.10543}.

\bibitem[Nathan~Silberman \& Fergus(2012)Nathan~Silberman and Fergus]{Silberman:ECCV12}
Pushmeet~Kohli Nathan~Silberman, Derek~Hoiem and Rob Fergus.
\newblock Indoor segmentation and support inference from rgbd images.
\newblock In \emph{ECCV}, 2012.

\bibitem[Oquab et~al.(2024)Oquab, Darcet, Moutakanni, Vo, Szafraniec, Khalidov, Fernandez, Haziza, Massa, El-Nouby, Assran, Ballas, Galuba, Howes, Huang, Li, Misra, Rabbat, Sharma, Synnaeve, Xu, Jegou, Mairal, Labatut, Joulin, and Bojanowski]{oquabDINOv2LearningRobust2024}
Maxime Oquab, Timothée Darcet, Théo Moutakanni, Huy Vo, Marc Szafraniec, Vasil Khalidov, Pierre Fernandez, Daniel Haziza, Francisco Massa, Alaaeldin El-Nouby, Mahmoud Assran, Nicolas Ballas, Wojciech Galuba, Russell Howes, Po-Yao Huang, Shang-Wen Li, Ishan Misra, Michael Rabbat, Vasu Sharma, Gabriel Synnaeve, Hu~Xu, Hervé Jegou, Julien Mairal, Patrick Labatut, Armand Joulin, and Piotr Bojanowski.
\newblock {DINOv2}: {Learning} {Robust} {Visual} {Features} without {Supervision}, February 2024.
\newblock URL \url{http://arxiv.org/abs/2304.07193}.
\newblock arXiv:2304.07193 [cs].

\bibitem[Radford et~al.(2021)Radford, Kim, Hallacy, Ramesh, Goh, Agarwal, Sastry, Askell, Mishkin, Clark, Krueger, and Sutskever]{radfordLearningTransferableVisual2021}
Alec Radford, Jong~Wook Kim, Chris Hallacy, Aditya Ramesh, Gabriel Goh, Sandhini Agarwal, Girish Sastry, Amanda Askell, Pamela Mishkin, Jack Clark, Gretchen Krueger, and Ilya Sutskever.
\newblock Learning {Transferable} {Visual} {Models} {From} {Natural} {Language} {Supervision}, February 2021.
\newblock URL \url{http://arxiv.org/abs/2103.00020}.
\newblock arXiv:2103.00020 [cs].

\bibitem[Ranftl et~al.(2021)Ranftl, Bochkovskiy, and Koltun]{ranftlVisionTransformersDense2021}
Rene Ranftl, Alexey Bochkovskiy, and Vladlen Koltun.
\newblock Vision {Transformers} for {Dense} {Prediction}.
\newblock In \emph{2021 {IEEE}/{CVF} {International} {Conference} on {Computer} {Vision} ({ICCV})}, pp.\  12159--12168, Montreal, QC, Canada, October 2021. IEEE.
\newblock ISBN 978-1-66542-812-5.
\newblock \doi{10.1109/ICCV48922.2021.01196}.
\newblock URL \url{https://ieeexplore.ieee.org/document/9711226/}.

\bibitem[Shi et~al.(2024)Shi, Liu, Wang, Liao, Radhakrishnan, Huang, Yin, Sapra, Yacoob, Shi, Catanzaro, Tao, Kautz, Yu, and Liu]{shi2024eagleexploringdesignspace}
Min Shi, Fuxiao Liu, Shihao Wang, Shijia Liao, Subhashree Radhakrishnan, De-An Huang, Hongxu Yin, Karan Sapra, Yaser Yacoob, Humphrey Shi, Bryan Catanzaro, Andrew Tao, Jan Kautz, Zhiding Yu, and Guilin Liu.
\newblock Eagle: Exploring the design space for multimodal llms with mixture of encoders, 2024.
\newblock URL \url{https://arxiv.org/abs/2408.15998}.

\bibitem[Steiner et~al.(2022)Steiner, Kolesnikov, Zhai, Wightman, Uszkoreit, and Beyer]{steiner2022trainvitdataaugmentation}
Andreas Steiner, Alexander Kolesnikov, Xiaohua Zhai, Ross Wightman, Jakob Uszkoreit, and Lucas Beyer.
\newblock How to train your vit? data, augmentation, and regularization in vision transformers, 2022.
\newblock URL \url{https://arxiv.org/abs/2106.10270}.

\bibitem[Tong et~al.(2024{\natexlab{a}})Tong, Brown, Wu, Woo, Middepogu, Akula, Yang, Yang, Iyer, Pan, Wang, Fergus, LeCun, and Xie]{tongCambrian1FullyOpen2024}
Shengbang Tong, Ellis Brown, Penghao Wu, Sanghyun Woo, Manoj Middepogu, Sai~Charitha Akula, Jihan Yang, Shusheng Yang, Adithya Iyer, Xichen Pan, Austin Wang, Rob Fergus, Yann LeCun, and Saining Xie.
\newblock Cambrian-1: {A} {Fully} {Open}, {Vision}-{Centric} {Exploration} of {Multimodal} {LLMs}, June 2024{\natexlab{a}}.
\newblock URL \url{http://arxiv.org/abs/2406.16860}.
\newblock arXiv:2406.16860 [cs].

\bibitem[Tong et~al.(2024{\natexlab{b}})Tong, Liu, Zhai, Ma, LeCun, and Xie]{tong2024eyes}
Shengbang Tong, Zhuang Liu, Yuexiang Zhai, Yi~Ma, Yann LeCun, and Saining Xie.
\newblock Eyes wide shut? exploring the visual shortcomings of multimodal llms.
\newblock In \emph{Proceedings of the IEEE/CVF Conference on Computer Vision and Pattern Recognition}, pp.\  9568--9578, 2024{\natexlab{b}}.

\bibitem[Xu et~al.(2019)Xu, Wang, Ceylan, Mech, and Neumann]{xu2019disn}
Qiangeng Xu, Weiyue Wang, Duygu Ceylan, Radomir Mech, and Ulrich Neumann.
\newblock Disn: Deep implicit surface network for high-quality single-view 3d reconstruction.
\newblock In \emph{NeurIPS}, 2019.

\bibitem[Yue et~al.(2024)Yue, Ni, Zhang, Zheng, Liu, Zhang, Stevens, Jiang, Ren, Sun, Wei, Yu, Yuan, Sun, Yin, Zheng, Yang, Liu, Huang, Sun, Su, and Chen]{yue2024mmmumassivemultidisciplinemultimodal}
Xiang Yue, Yuansheng Ni, Kai Zhang, Tianyu Zheng, Ruoqi Liu, Ge~Zhang, Samuel Stevens, Dongfu Jiang, Weiming Ren, Yuxuan Sun, Cong Wei, Botao Yu, Ruibin Yuan, Renliang Sun, Ming Yin, Boyuan Zheng, Zhenzhu Yang, Yibo Liu, Wenhao Huang, Huan Sun, Yu~Su, and Wenhu Chen.
\newblock Mmmu: A massive multi-discipline multimodal understanding and reasoning benchmark for expert agi, 2024.
\newblock URL \url{https://arxiv.org/abs/2311.16502}.

\bibitem[Zhai et~al.(2023)Zhai, Mustafa, Kolesnikov, and Beyer]{zhaiSigmoidLossLanguage2023}
Xiaohua Zhai, Basil Mustafa, Alexander Kolesnikov, and Lucas Beyer.
\newblock Sigmoid {Loss} for {Language} {Image} {Pre}-{Training}, September 2023.
\newblock URL \url{http://arxiv.org/abs/2303.15343}.
\newblock arXiv:2303.15343 [cs].

\bibitem[Zhan et~al.(2024)Zhan, Zheng, Xie, and Zisserman]{zhanGeneralProtocolProbe2024}
Guanqi Zhan, Chuanxia Zheng, Weidi Xie, and Andrew Zisserman.
\newblock A {General} {Protocol} to {Probe} {Large} {Vision} {Models} for {3D} {Physical} {Understanding}, June 2024.
\newblock URL \url{http://arxiv.org/abs/2310.06836}.
\newblock arXiv:2310.06836 [cs].

\end{thebibliography}
\bibliographystyle{colm2025_conference}

\clearpage
\setcounter{page}{1}
\appendix
\section{Overview of task examples}
\begin{figure}[h]
    \centering
    \includegraphics[width=1\linewidth]{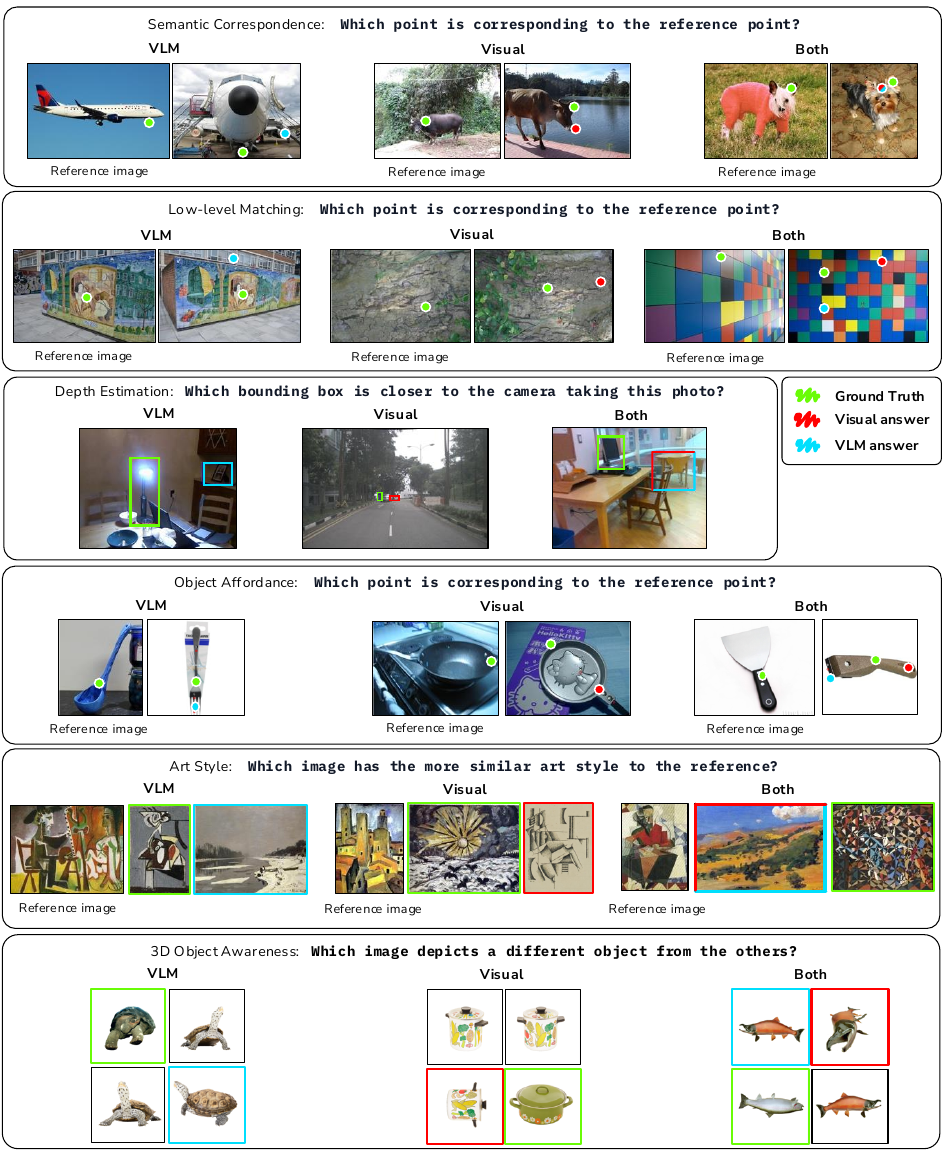}
    \caption
    {\small \textbf{Example failure cases for VLMs (left), vision encoders (center), and both (right).} We observe a few common failure modes for
both evaluation strategies: on correspondence-based tasks, similar local structure may confuse the model (see Low-level Matching: stone
wall and colorful squares). Additionally, some objects or object parts of interest may be too small to be individually encoded in high fidelity
(see Semantic Correspondence: dog ear, and Depth Estimation: small vehicles).
    }
    \label{fig:failures}
\end{figure}

\section{Multi-image VLM inputs}
\label{sec:multi}

\subsection{Evaluation setup}
\label{subsec:multi_setup}
Certain VLM architectures have the native ability to handle multi-image inputs without needing to stitch them into a single input.
As checkpoints for these architectures are not released with multiple visual encoder backbones, they are not appropriate for the direct comparison methods provided in the main paper. However, they offer insights into the upper bound of current VLM performance on some vision centric tasks and remove any possible failure modes associated with the image-stitching method. We focus on the performance for 3D Object Awareness task.  
The models we examine are LLaVA-Next \cite{liu2024llavanext}, Phi-3 \cite{abdin2024phi3technicalreporthighly}, and Qwen-VL \cite{bai2023qwenvlversatilevisionlanguagemodel}. We follow the multi-image prompting methods of each codebase. The following prompt is used for all of the VLMs, with \texttt{n} modulated for the correct number of image inputs:

\noindent \texttt{Given n images, all but one of them depicts the same object from a different angle. Can you tell which image is the odd one out? Select from the following choices. (A) the first image (B) the second image (C) the third image \\
Answer with the option's letter from the given choices directly.}

\subsection{Results}
\label{subsec:multi_results}
\begin{table}[h!]
\centering
\begin{tabular}{@{}c@{ }c@{ }c@{ }c@{}}
\toprule
LLaVA-Next & Phi-3 & Qwen-VL & \textbf{Qwen-VL 1Shot} \\ \midrule
$0.340 \pm 0.02$ & $0.320 \pm 0.02$ & $0.430 \pm 0.02$ &  $0.452 \pm 0.02$\\
\bottomrule
\end{tabular}
\caption{\textbf{Performance of LLaVA-Next, Phi-3, Qwen-VL, Qwen-VL 1 Shot models on the 3D Object Awareness task. } We find that Qwen-VL performs the best and significantly above chance, and improves slightly with one in-context example. However, these results do not account for the visual-VLM gap observed in the main paper's results.
}
\end{table}
\label{tab:vlm_new_results}

As shown in Table \ref{tab:vlm_new_results}, we see that only one model, Qwen-VL, performs considerably better than random chance ($=0.325$) across the tasks and models explored.
Furthermore we see that performance continues to improve when in-context learning is used to extend the best performing approach to the 1-shot setting. These results show that the state-of-the-art open-source VLMs, when given proper multi-image inputs, can perform above random chance but still do not account for the gap between VLM and visual evaluation strategies.

\section{Evaluation details}
\label{sec:eval_details}

For all our evaluations in the main paper, we use four VLMs from the suite of models provided by Prismatic VLMs \cite{karamchetiPrismaticVLMsInvestigating2024}. These models have corresponding vision backbones found in the \texttt{timm} library (except the CLIP vision model, which we source from the \texttt{openclip} library): 

\begin{table}[h]
\centering
\scalebox{0.88}{
\begin{tabular}{@{}rl@{}}
\toprule
\multicolumn{1}{c}{\textbf{VLM (Prismatic)}} & \multicolumn{1}{c}{\textbf{Vision backbone (timm and openclip)}} \\ \midrule
\texttt{in1k-224px+7b} & \texttt{vit\_large\_patch16\_224} \\
\texttt{dinov2-224px+7b} & \texttt{vit\_large\_patch16\_224} \\
\texttt{clip-224px+7b} & \texttt{ViT-L-14/openai} \\
\texttt{siglip-224px+7b} & \texttt{vit\_large\_patch16\_siglip\_256} \\ \bottomrule
\end{tabular}
}
\end{table}

Each VLM is made up of a vision backbone, Vicuña v1.5 \cite{vicuna2023} as the LM backbone, and a projector module (2-layer MLP). The vision and language components are frozen, while the projector is trained on the LLaVA v1.5 pre-training set (refer to \cite{karamchetiPrismaticVLMsInvestigating2024}'s single-stage training procedure).

We preprocess VLM image inputs with the model's default `letterbox padding' scheme to the desired dimensions (specified in the task descriptions below), which is the same procedure as performed by LLaVA v1.5 \cite{liuImprovedBaselinesVisual2023}. For the vision encoder-only experiments, we perform a naïve resize to the desired dimensions. Normalization is performed with the same mean and std for each pair of models, and corresponds to the original model specifications (e.g., ImageNet statistics for DINOv2). We use some basic `fuzzy-matching' to extract the letter choice from the VLM, allowing for some variance in output format.

Below, we describe details for each evaluation from the main paper. 

\subsection{Depth Estimation}

For each sample, we resize the images to have a height of 1024px and use the prompts provided by \cite{tongCambrian1FullyOpen2024}, which are of the following format:

\noindent \texttt{Which object is closer to the camera taking this photo, the table (highlighted by a red box) or the bookcase (highlighted by a blue box)? (A) table (B) bookcase \\
Answer with the option's letter from the given choices directly.}

To evaluate vision encoders, we train a DPT decoder following \cite{bananiProbing3DAwareness2024} and apply it to 4 layers uniformly spaced throughout the model. Specifically, we initialize the decoder architecture introduced in \cite{ranftlVisionTransformersDense2021} and train on the NYUv2 depth training set \cite{Silberman:ECCV12} for 10 epochs with the AdamW optimizer (\texttt{lr}$=5\text{e-}4$, and 1.5 epochs of linear warmup with cosine decay). 

Given a predicted depth map at inference time, we crop to the two query bounding boxes and average the depths in each box (effectively getting an average depth of each object in question). We acknowledge that the bounding box, without any proper segmentation mask, also captures pixels not associated with the object (e.g., space in-between chair legs or next to a tilted object), and that this simple evaluation method may actually under-estimate the performance of the visual readout strategy. The final prediction from the vision model comes from a comparison of these two scalar depth values.

We construct further examples to finetune the VLM by randomly sampling images from Omni3D \cite{brazil2023omni3dlargebenchmarkmodel} and drawing bounding boxes over two random labeled objects in the scene.

\subsection{Correspondence-based Evaluations}
The following details apply to the Semantic Correspondence, Object Affordance, and Low-Level Matching benchmarks, which all follow the same task format.

Our VLM evaluations use the prompts provided by \cite{fuBLINKMultimodalLarge2024}, which are of the following format (this one taken from the Semantic Correspondence task):

\noindent \texttt{Humans can find corresponding points for different objects in the same category. For instance, if there are images of two different cats, then the left ear tip of one cat corresponds to the left ear tip of the other cat, and the right front paw of one cat corresponds to the right front paw of the other cat. Given the following two images, a reference point is annotated on the first image, labeled with REF. You are given multiple red-circled points on the second image, choices of "A, B, C, D" are drawn beside each circle. Select between the choices on the second image and find the corresponding point for the reference point. Which point is corresponding to the reference point? Select from the following choices. (A) Point A (B) Point B (C) Point C (D) Point D \\
Answer with the option's letter from the given choices directly.}

We evaluate vision encoders first by locating the coordinates of `REF', `A', `B', `C', and `D' by visually matching the text and red circles in the query image. Then, we extract the last-layer patch features of the ViT and sample floating-point coordinates of each red-circled point (with bilinear interpolation). We then compute cosine similarity between the `REF' point and all other points, choosing the option with the highest similarity value as the final answer. For both evaluation methods, we resize images to a height of 224px. This lower resolution is required for VLMs to fit the stitched input image into its context length.

We construct further examples to finetune the VLM by randomly sampling images from the same datasets (SPair-71k, HPatches, and FunKPoint). We sample a reference keypoint and four multiple choice options, superimposing the same red circles and labels as in the evaluation benchmarks.
\subsection{3D Object Awareness}

We adapt the MOCHI benchmark \cite{bonnen2024evaluatingmultiviewobjectconsistency} to be a VLM task with the following prompt, where \texttt{n} is 3 or 4:

\noindent \texttt{Given n images, all but one of them depicts the same object from a different angle. Can you tell which image is the odd one out? Select from the following choices. (A) the first image (B) the second image (C) the third image \\
Answer with the option's letter from the given choices directly.}

In cases with 4 images, we add \texttt{(D) the fourth image} as an option in the prompt.

Our vision evaluation follows that of \cite{bonnen2024evaluatingmultiviewobjectconsistency}, where the global \texttt{CLS} token is extracted from the vision encoder and a pairwise cosine similarity matrix is computed. As the final model prediction, we select the image with the lowest average similarity in the matrix. For both evaluation methods, we resize images to a height of 224px. This lower resolution is required for VLMs to fit the stitched input image into its context length.

To finetune VLMs for this task, we sample ShapeNet renderings from \cite{xu2019disn, chang2015shapenetinformationrich3dmodel} and pose a 3-way multiple choice task.

\subsection{Art Style}

We evaluate VLMs with the following prompt, taken from the BLINK benchmark \cite{fuBLINKMultimodalLarge2024}:

\noindent \texttt{Some most common art painting styles include Realism, Impressionism, Expressionism, Pop Art, and Cubism. Given the following images of art paintings, use the first image as the reference image, and determine which one of the second or the third image shares the same style as the reference image? Select from the following choices. (A) the second image (B) the third image \\
Answer with the option's letter from the given choices directly.}

We evaluate vision encoders by computing the Gram matrices of the ViT's patch features. Specifically, we obtain patch features $F$ for each image of dimension $C \text{ x } HW$, and multiply $FF^T$ to get a matrix $G$ with spatial information removed, preserving mostly texture, or `style' information.

While this method of obtaining a `style matrix' for ViTs has been found to give blocky style-transferred outputs and a smoother solution has been found by \cite{10.1145/3512353.3512366}, we find that this method still captures the texture-focused style information required for the task and keep this simpler evaluation rather than a method tailored for high-quality style transfer results. 

\begin{figure}[]
    \centering
    \includegraphics[width=0.5\linewidth]{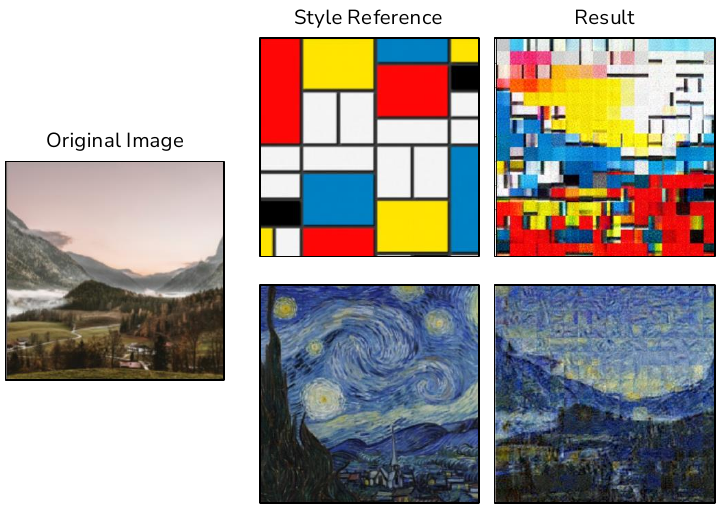}
    \caption
    {Two examples of `neural style transfer': starting with the original image, we adjust the pixels by minimizing mean squared error between $G_{\text{original image}}$ and $G_{\text{style reference}}$. The resulting image takes on the texture of the middle reference image, while retaining the high-level structure of the landscape image.
    }
    \label{fig:style_sanity}
\end{figure}

See Figure \ref{fig:style_sanity} for examples of a simplified version of \cite{Gatys2015ANA}'s neural style transfer, where we optimize the original input image to minimize $MSE(G_{\text{result}}, G_{\text{style reference}})$. For both evaluation methods, we resize images to a height of 224px. This lower resolution is required for VLMs to fit the stitched input image into its context length.

To finetune VLMs for the Art Style task, we sample images from WikiArt and create positive pairs from two images with the same art style label. The negative example is randomly chosen from all other art styles.

\subsection{Few-shot evaluation details}
We use a logistic regression classifier on \texttt{CLS} embeddings to evaluate a vision model $f$ on the 3D Object Awareness task in a few-shot setting. Given 3 images $\{x_0, x_1, x_2\}$, we compute pairwise differences $| f(x_0) - f(x_1) |$ between images and label each difference vector with a 1 if it contains the odd-one-out (0 otherwise). Then to perform inference for one data point belonging to some condition, we train a logistic regression classifier on 75\% of that condition (where condition subsets are defined in  \cite{bonnen2024evaluatingmultiviewobjectconsistency}) and evaluate on the single data point. We repeat this procedure 10 times for each data point and report the mean and standard deviation (which is zero) in the main paper.

To evaluate VLMs, we tune one prompt embedding. Using the same train/test split procedure for each condition as described above, we concatenate one randomly-initialized embedding to the start of the prompt sequence (after the $<$BOS$>$ embedding) and train it on 75\% of the condition. We run this optimization for 10 trials for each data point and report the mean and standard deviation in the main paper. We use the Adam optimizer with a learning rate of 3e-3 and train for 10 epochs.

\newpage
\subsection{Additional attention visualizations}
\begin{figure*}[h]
    \centering
\includegraphics[width=0.71\linewidth]{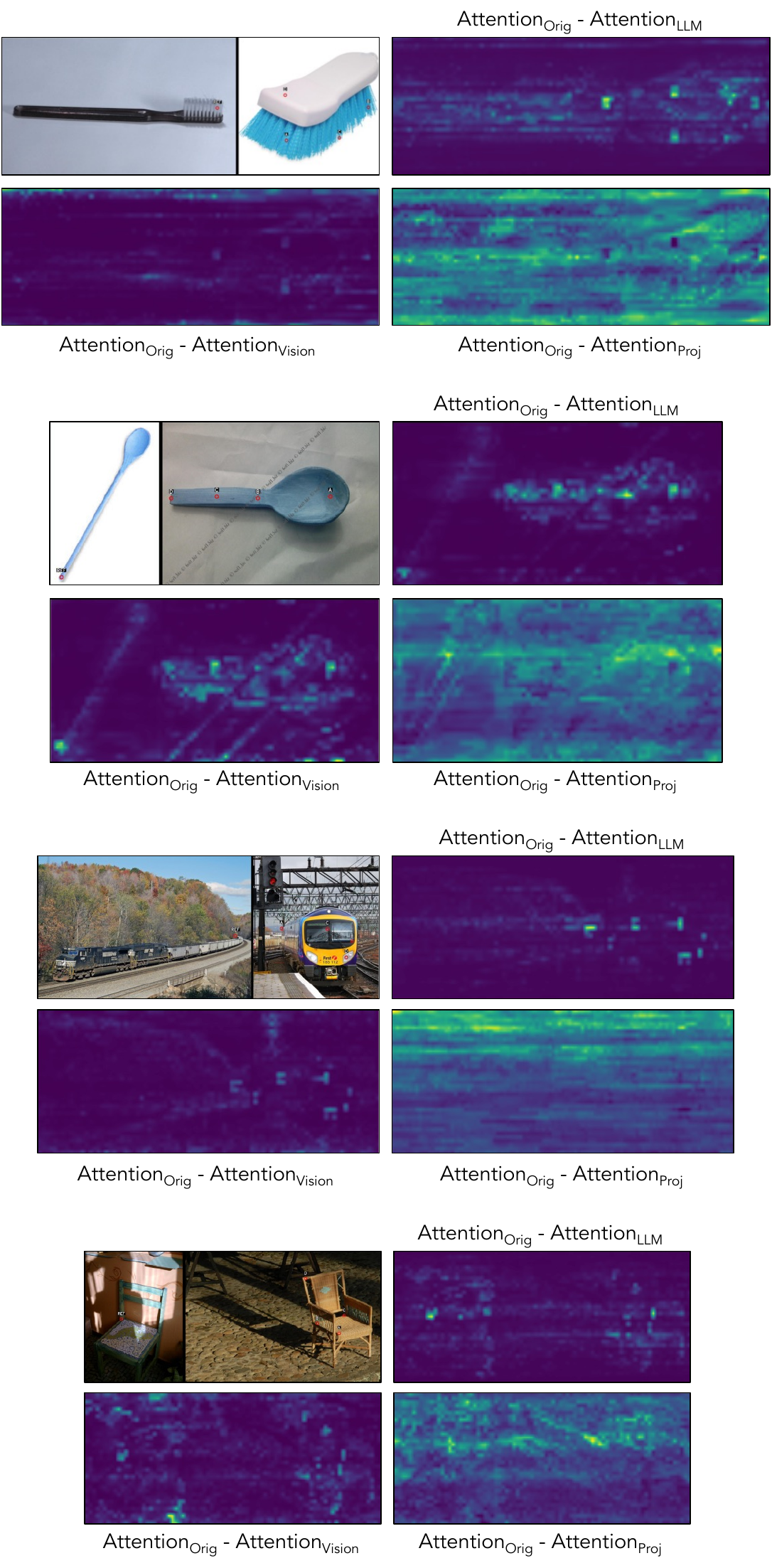}
\vspace{-2mm}
    \caption
    {Attention difference visualizations for Functional Correspondence (top two) and Semantic Correspondence (bottom two) tasks.} 
    \label{fig:attn_supp}
\end{figure*}

\section{Additional failure examples}
\begin{figure*}[h]
    \centering
    \includegraphics[width=1\linewidth]{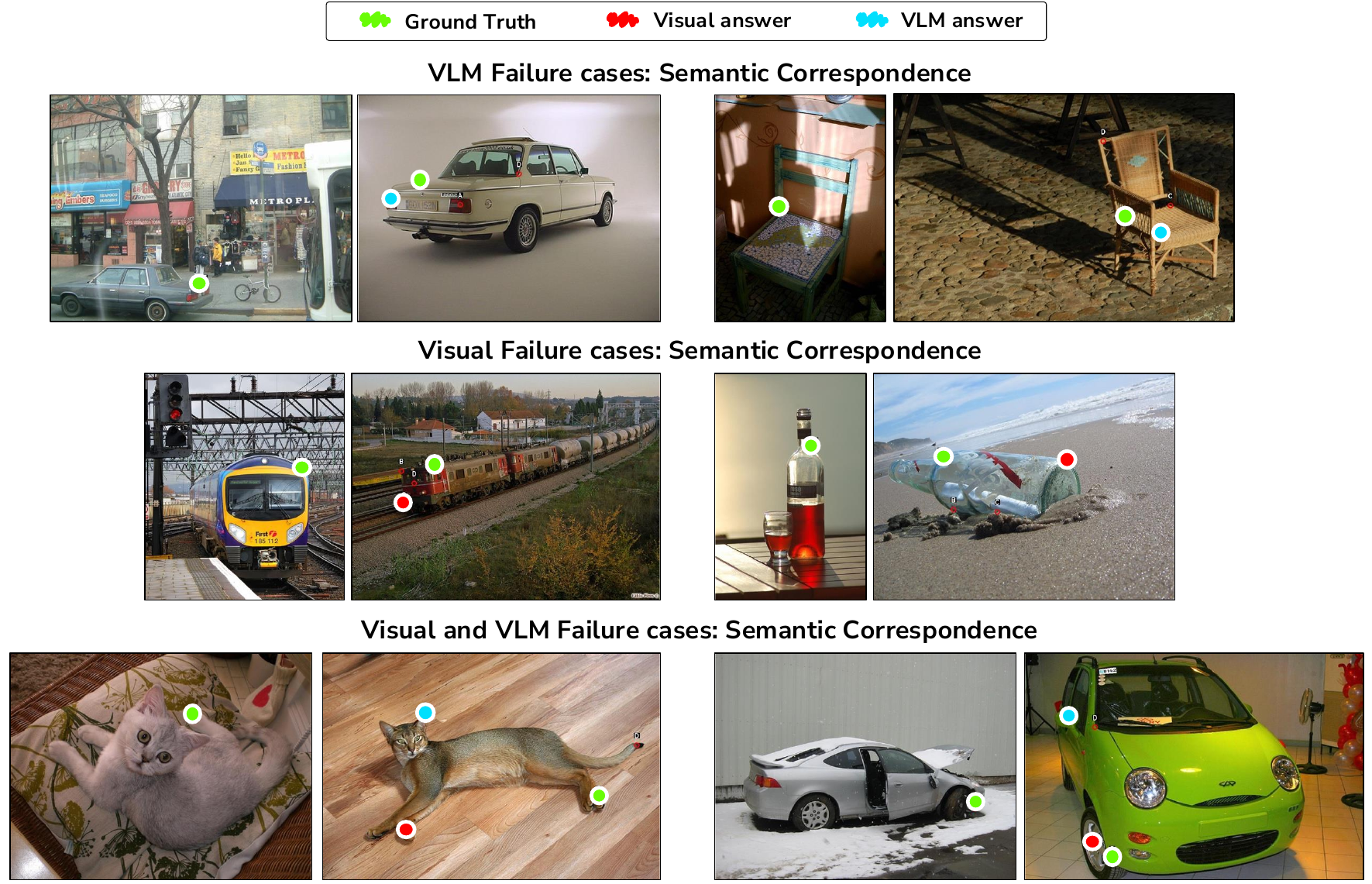}
    \caption
    {\textbf{Which point in the image (right) best matches the semantics of the point in the reference image (left)?} Additional examples of VLM-only, vision encoder-only, and VLM + vision encoder failures on the Semantic Correspondence task. }
    \label{fig:supp_semcorresp}
\end{figure*}

\begin{figure*}[h]
    \centering
    \includegraphics[width=1\linewidth]{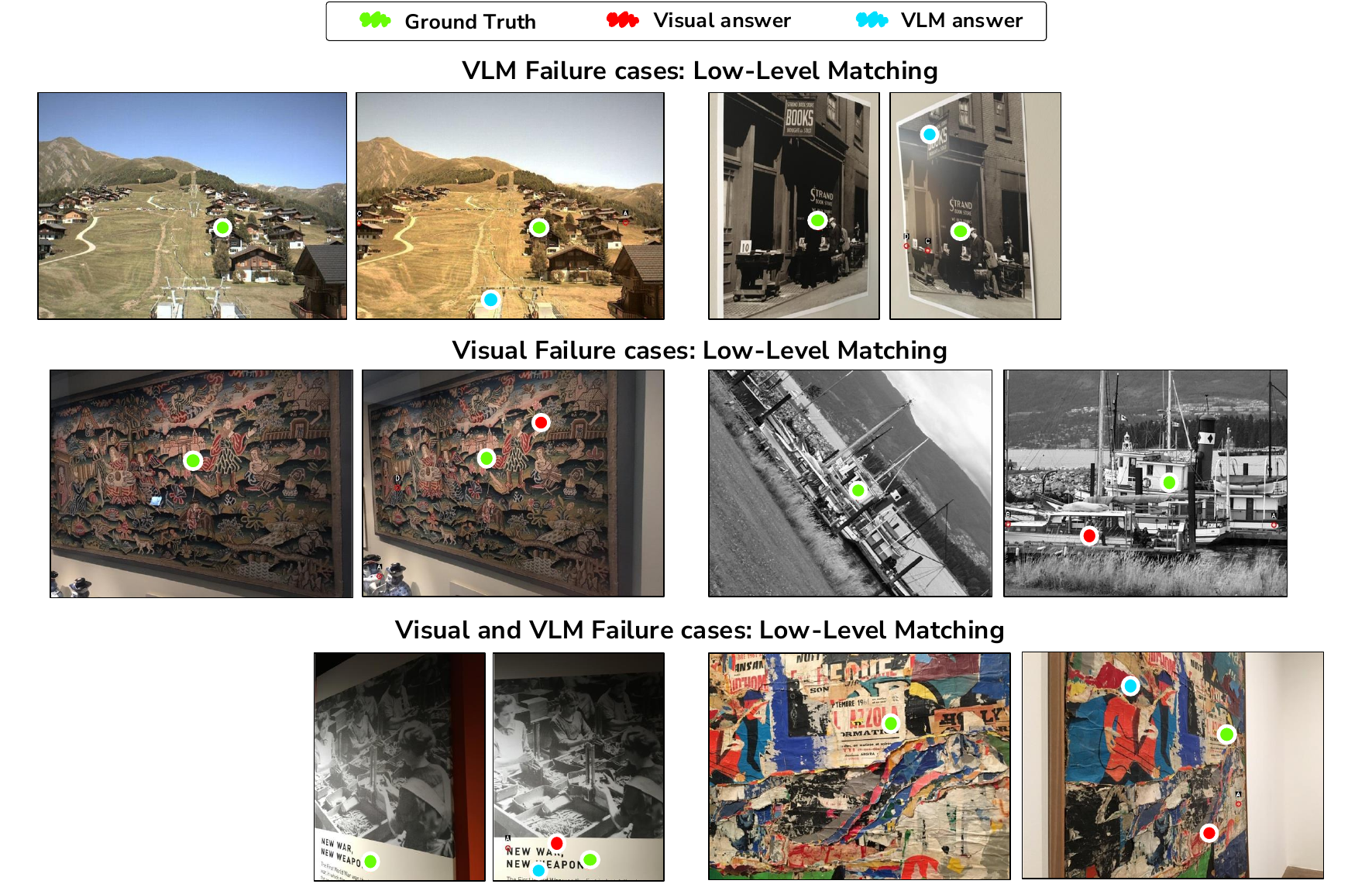}
    \caption
    {\textbf{Which point in the image (right) best matches to the reference point in the same scene (left)?} Additional examples of VLM-only, vision encoder-only, and VLM + vision encoder failures on the Low-level Matching task. }
    \label{fig:supp_llmatch}
\end{figure*}

\begin{figure*}[h]
    \centering
    \includegraphics[width=0.8\linewidth]{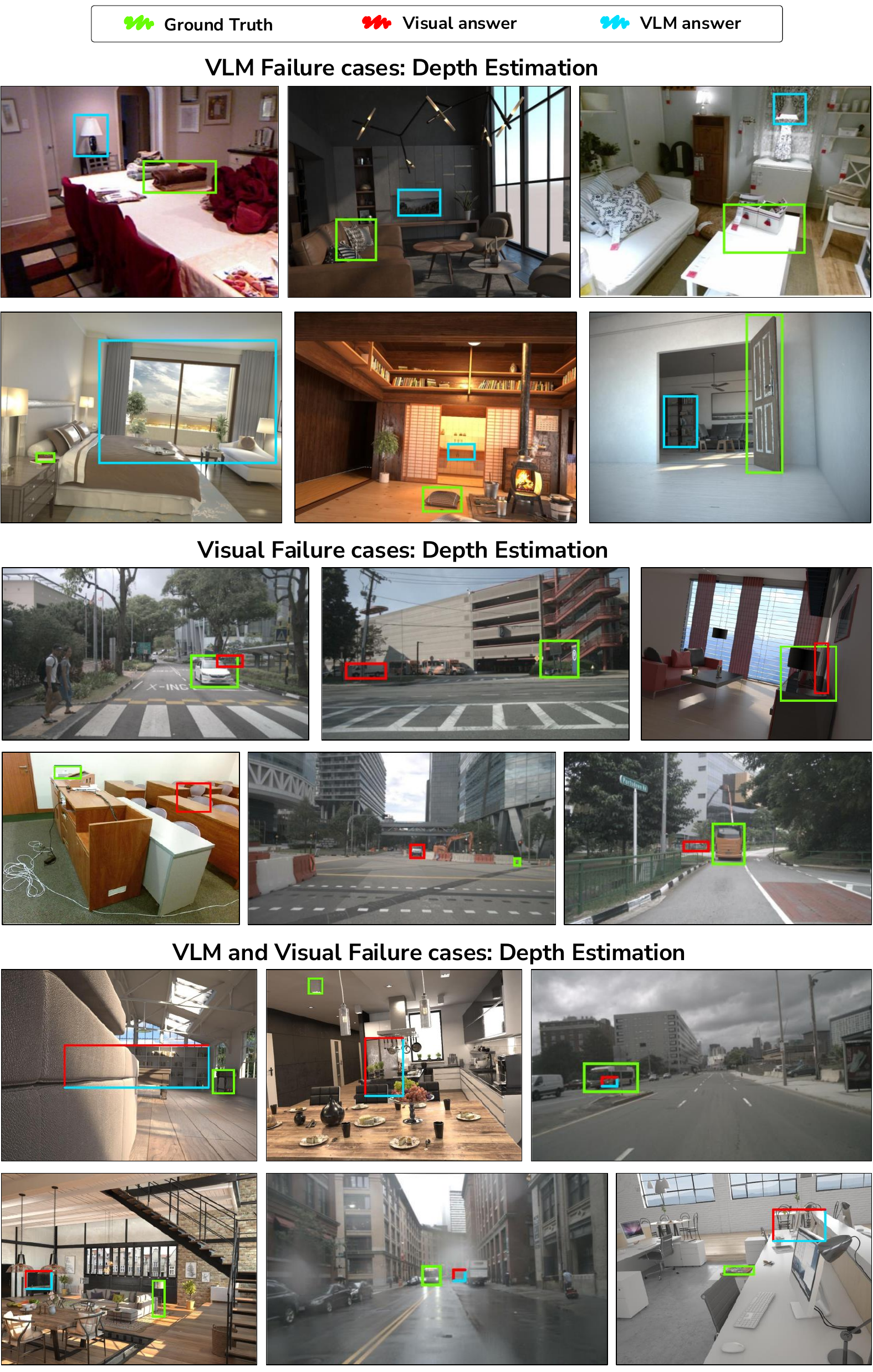}
    \caption
    {\textbf{Which bounding box contains the object closer to the camera? }Additional examples of VLM-only, vision encoder-only, and VLM + vision encoder failures on the Depth Estimation task. }
    \label{fig:supp_depth}
\end{figure*}

\begin{figure*}[h]
    \centering
    \includegraphics[width=1\linewidth]{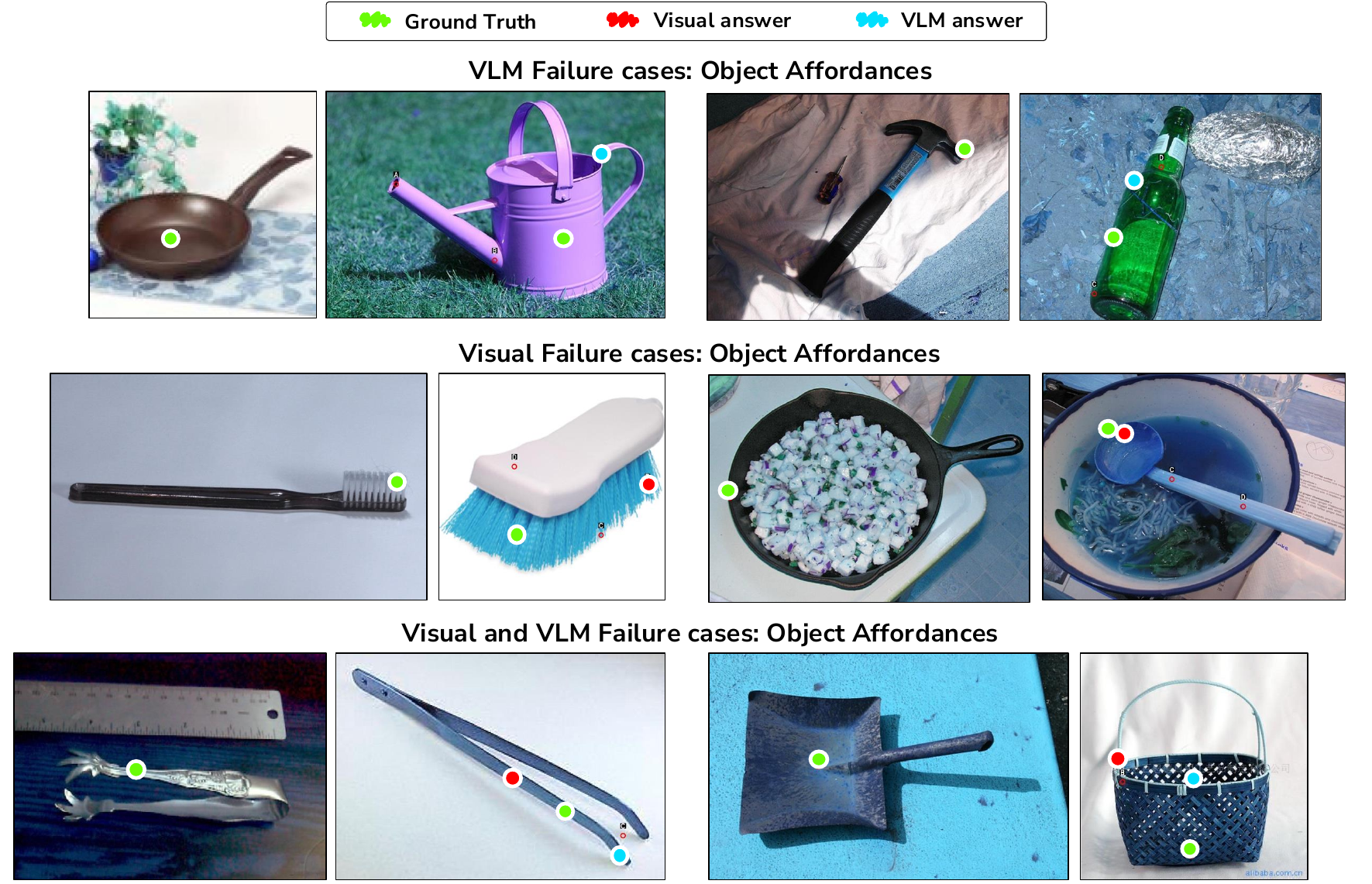}
    \caption
    {\textbf{Which point in the image (right) best matches the function of the point in the reference image (left)?} Additional examples of VLM-only, vision encoder-only, and VLM + vision encoder failures on the Object Affordances task. }
    \label{fig:supp_objaff}
\end{figure*}

\begin{figure*}[h]
    \centering
    \includegraphics[width=1\linewidth]{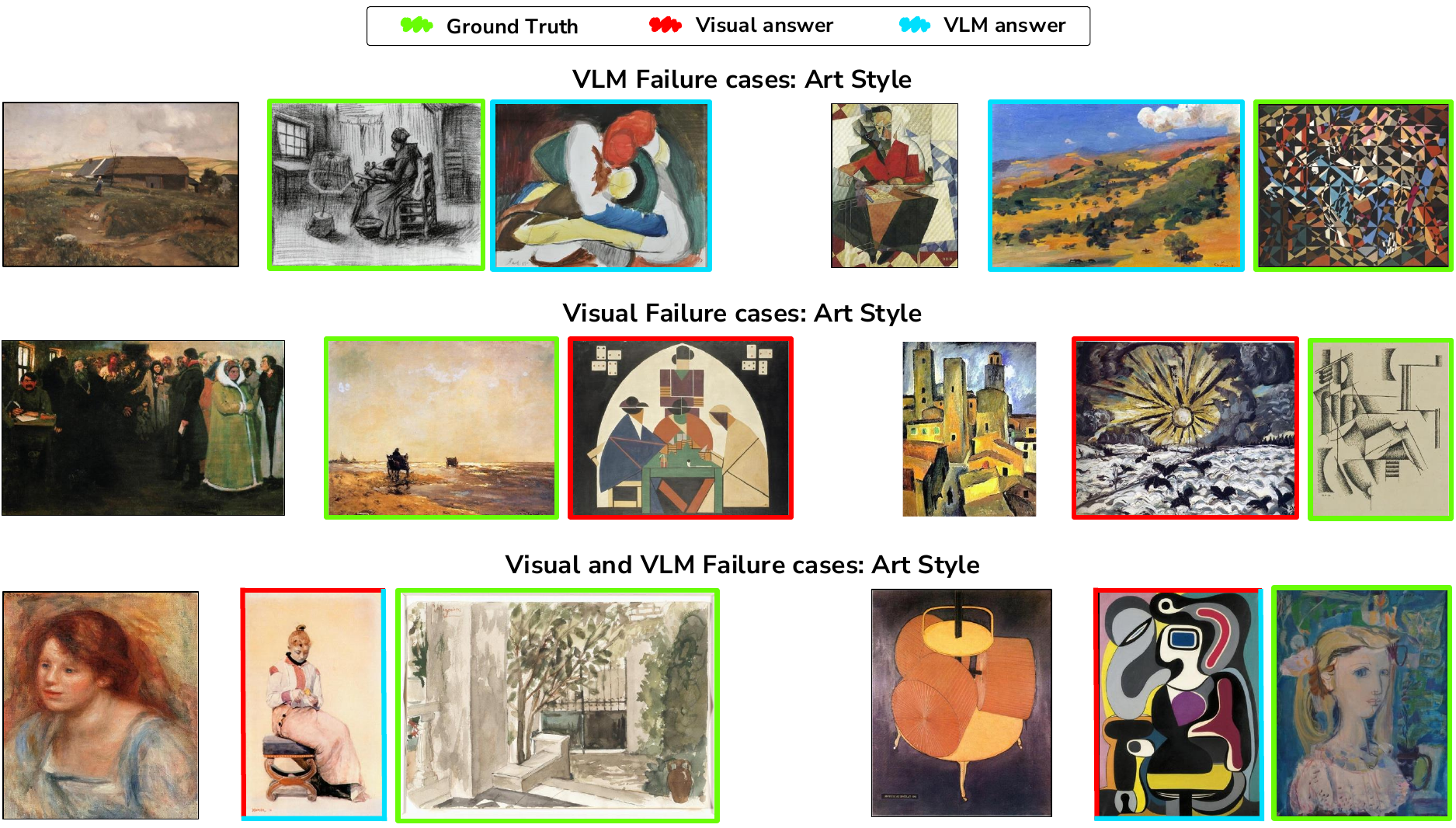}
    \caption
    {\textbf{Which images (center or right) best matches the art style of the reference image (left)? } Additional examples of VLM-only, vision encoder-only, and VLM + vision encoder failures on the Art Style task. }
    \label{fig:supp_artstyle}
\end{figure*}

\begin{figure*}[h]
    \centering
    \includegraphics[width=1\linewidth]{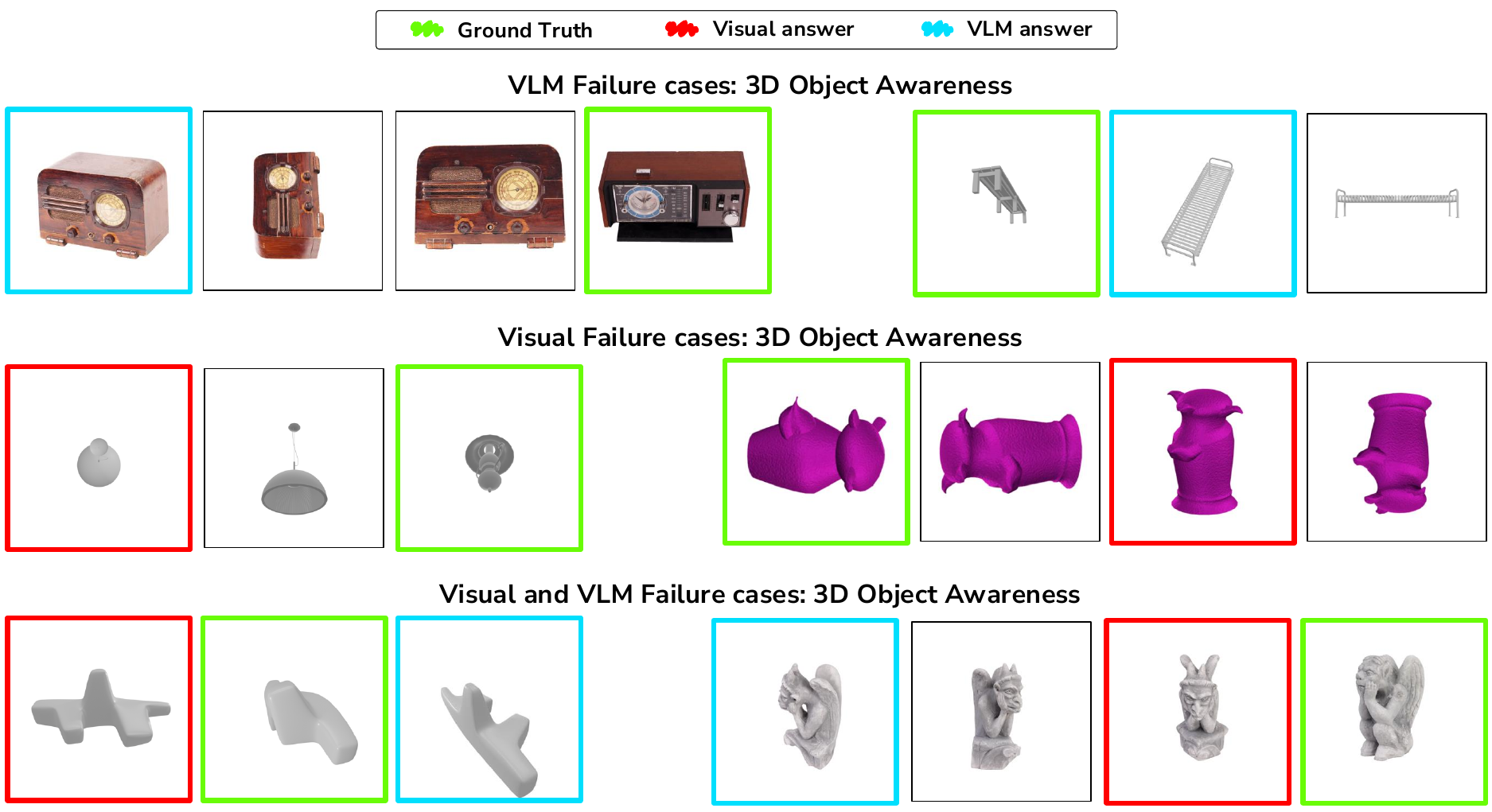}
    \caption
    {\textbf{Which image contains the odd-object-out? } Additional examples of VLM-only, vision encoder-only, and VLM + vision encoder failures on the 3D Object Awareness task. }
    \label{fig:supp_3dobj}
\end{figure*}

In Figures \ref{fig:supp_semcorresp}, \ref{fig:supp_llmatch}, \ref{fig:supp_depth}, \ref{fig:supp_objaff}, \ref{fig:supp_artstyle}, and \ref{fig:supp_3dobj}, we show additional non-cherrypicked examples from each benchmark where only the VLM fails, only the vision encoder fails, or both fail. Examples are chosen from models with DINOv2 as the vision backbone.

\end{document}